\documentclass[10pt,twocolumn,letterpaper]{article}

\usepackage{iccv} 

\newcommand{\TODO}[1]{\textbf{\color{red}[TODO: #1]}}
\renewcommand{\TODO}[1]{}

\usepackage{algpseudocode}
\usepackage{algorithm}
\usepackage{amsmath}
\usepackage{wrapfig}
\usepackage{amsthm}
\usepackage{diagbox}
\usepackage{amsfonts}
\usepackage[utf8]{inputenc} 
\usepackage[T1]{fontenc}    
\usepackage{url}            
\usepackage{booktabs}       
\usepackage{amsfonts}       
\usepackage{nicefrac}       
\usepackage{microtype}      
\usepackage{xcolor}         
\usepackage{bbm}
\usepackage{graphicx}
\newtheorem{theorem}{Theorem}

\newtheorem{proposition}{Proposition}
\newtheorem{definition}{Definition}
\newtheorem{lemma}{Lemma}

\usepackage{tabularray}
\usepackage{array}
\usepackage{pifont}

\definecolor{iccvblue}{rgb}{0.21,0.49,0.74}
\definecolor{mygreen}{RGB}{145,183,140}
\definecolor{myred}{RGB}{171, 90, 83}
\usepackage[nodisplayskipstretch]{setspace}
\usepackage[pagebackref,breaklinks,colorlinks,allcolors=iccvblue]{hyperref}

\def\paperID{13078} 
\def\confName{ICCV}
\def\confYear{2025}

\title{Sharpness-Aware Teleportation on Riemannian Manifolds}

\author{Tuan Truong\\
Qualcomm AI Research\\
Vietnam\\
{\tt\small tuantruo@qti.qualcomm.com}
\and
Hoang-Phi Nguyen\\
Qualcomm AI Research\\
Vietnam\\
{\tt\small phinguye@qti.qualcomm.com}
\and
Haocheng Luo\\
Monash University\\
Australia\\
{\tt\small haocheng.luo@monash.edu}
\and
Tung Pham\\
Qualcomm AI Research\\
Vietnam\\
{\tt\small tungp@qti.qualcomm.com}
\and
Mehrtash Harandi\\
Monash University\\
Australia\\
{\tt\small Mehrtash.Harandi@monash.edu}
\and
Dinh Phung\\
Monash University\\
Australia\\
{\tt\small Dinh.Phung@monash.edu}
\and
Trung Le\\
Monash University\\
Australia\\
{\tt\small trunglm@monash.edu}
}

\begin{document}
\maketitle

\begin{abstract}
Recent studies highlight the effectiveness of flat minima in enhancing generalization, with sharpness-aware minimization (SAM) achieving state-of-the-art performance. Additionally, insights into the intrinsic geometry of the loss landscape have shown promise for improving model generalization. Building on these advancements, we introduce a novel sharpness-aware, geometry-aware teleportation mechanism to further enhance robustness and generalization. The core innovation of our approach is to decompose each iteration into a teleportation step within a local orbit and a sharpness-aware step that transitions between different orbits, leveraging the Riemannian quotient manifold. Our approach is grounded in a theoretical framework that analyzes the generalization gap between population loss and worst-case empirical loss within the context of Riemannian manifolds. To demonstrate the effectiveness of our method, we evaluate and compare our algorithm on diverse vision benchmarks with various datasets and Riemannian manifolds.
\end{abstract}

\section{Introduction}

A primary challenge with deep learning models is overfitting, where the model struggles to generalize because it gets trapped in local minima within the loss landscape during training. Overfitting affects various computer vision applications; for instance, in object detection, a model may rely on background context cues rather than the object itself, failing to generalize to different backgrounds \citep{torralba2010context}. Similarly, in semantic segmentation, models often overfit to specific textures and patterns within the training set, which may not be consistent in new data \citep{chen2018deeplab}. This issue stems from the high-dimensional, non-convex nature of the loss landscape, which contains numerous local minima. Identifying "good" local minima, often associated with "flat" or "low-sharpness" regions, has been shown to improve generalization. Sharpness-aware minimization (SAM) \citep{sam} is a notable approach to this problem, as it minimizes both the model’s loss and the worst-case loss within a neighborhood of the parameter space. SAM has demonstrated effectiveness across diverse areas, including meta-learning \citep{abbas2022sharp}, federated learning \citep{qu2022generalized}, vision models \citep{chen2021vision}, and language models \citep{bahri-etal-2022-sharpness}.

In addition to generalization, model robustness is a highly desirable property, particularly in computer vision. For instance, in medical imaging, variations in patient demographics—such as differences in anatomy and skin tone—can lead to misdiagnoses if a model lacks robustness \citep{gulshan2016development}. To enhance robustness, prior works often incorporate constraints on parameters, such as symmetric positive definiteness (SPD) \citep{spd}, full rank, or orthogonality \citep{orthocnn, siamesemanifold, wang2020orthogonal}. These strict constraints increase the complexity of training, as the optimization must be conducted on a Riemannian manifold rather than in Euclidean space, introducing a more complex geometric structure to the optimization process. Consequently, a range of Riemannian optimization methods, including Riemannian stochastic gradient descent (RSGD), Riemannian gradient descent (RGD), Riemannian accelerated SGD (RASGD), and others, have been developed to address these challenges \citep{rsgd, rgd, rasgd, rsvg, rsvrg}.

In machine learning, and particularly in computer vision, group-invariance properties have been extensively explored. For example, Capsule Networks \citep{sabour2017dynamic} were designed to achieve rotational and translational invariance through dynamic routing, while models like Feature Pyramid Networks (FPN) and YOLO employ multi-scale processing to achieve scale invariance. As group-invariant architectures advance, developing optimization techniques that effectively leverage these invariant properties offers a promising approach for more efficient optimization. \citep{siamesemanifold} suggests that instead of optimizing over the entire parameter space, it is more efficient to restrict the search to the quotient space under group actions, which typically has lower dimensionality. More recently, \citep{teleport2} and \citep{teleportoriginal} introduced the concept of \textit{teleportation}, which identifies the group action that maximizes the gradient norm at each iteration, allowing the model to "teleport" to a new parameter. This approach aims to maximize step size at each iteration and accelerate the optimization process.  

In this work, we propose Riemannian Sharpness-Aware Teleportation (RSAT), a novel Riemannian optimizer designed to leverage the group-invariant structure of the parameter space to enhance both generalization and model robustness on Riemannian manifolds. RSAT divides each optimization iteration into two distinct phases. The first phase focuses on finding the optimal group action that minimizes the gradient norm and \textbf{teleport within the orbit}. This phase adjusts the sharpness without affecting the loss value, effectively refining the model's robustness. The second phase performs sharpness-aware minimization on the Riemannian quotient manifold, \textbf{moves between different orbits} and thereby reduces the search space to the lower-dimensional quotient space. Together, these two phases work synergistically to minimize sharpness, improving generalization and robustness within the Riemannian context.

The development of RSAT is underpinned by a theoretical analysis that characterizes the generalization gap in the Riemannian setting, considering the group-invariance structure of the loss function. Our empirical results demonstrate that RSAT notably improves generalization across a variety of tasks, datasets, and architectures. In particular, RSAT outperforms existing methods, including Sharpness-Aware Minimization (SAM) \citep{sam}, Supervised Contrastive Learning (SupCon) \citep{supcon}, and other Riemannian optimizers such as Riemannian Stochastic Gradient Descent (RSGD) \citep{rsgd}, Riemannian Adam (R-Adam) \cite{radam},  and Riemannian-SAM \citep{rsam}. In summary, our contributions are:

\begin{itemize}
    \item We introduce RSAT, a Riemannian optimizer with a novel teleportation mechanism designed to improve generalization. This algorithm is grounded on a theoretical analysis showing the generalization gap on Riemannian manifolds. 
    \item  We demonstrate the effectiveness of RSAT on various datasets and manifolds, showing that it outperforms baselines such as SAM \citep{sam}, RSGD \citep{rsgd}, Symmetry Teleportation \citep{teleportoriginal}, R-Adam \cite{radam}, and Riemannian-SAM \citep{rsam}.
\end{itemize}

\begin{figure}
    \centering
    \includegraphics[width=0.35\textwidth]{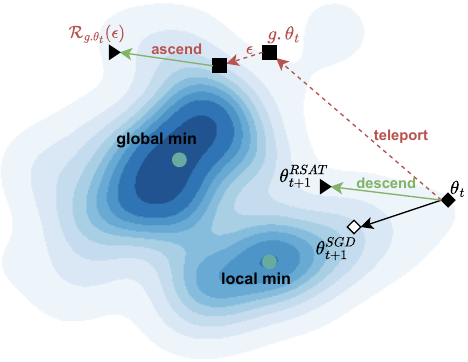} 
    \caption{Schematics of RSAT {\color{mygreen} (green)} vs. SGD (black) parameter update. RSAT "teleports" in the same orbit {\color{myred}(red)}, ascends to minimize the sharpness and descends to another orbit.}
    \label{fig: schematic}
    \vspace{-\baselineskip}
\end{figure}


\section{Preliminaries}
\subsection{Notations and formulation}
In this section, we introduce the problem's formulation and necessary notations to prepare for the later parts of this work. Assume that we are given a data set, says $\mathcal{S}$, consisting of many pairs of $\{(x_i,y_i): i=1,2,\ldots,n\}$, in which $x_i$ is a datum and $y_i$ is the $x_i$'s label, i.e. $x$ is an image of a dog and $y= \textrm{``dog"}$. Assume that $x$ is recorded as a point in $\mathbb{R}^k$ and $y$ is encoded to a number among $\{1,\ldots,K\}$. The task is to build a classifier that could assign datum $x\in \mathbb{R}^k$ to its correct label $y$. We parameterize the classifier by a function  $f_{\mathbf{\theta}}$ with $\mathbf{\theta}$ as a hyperparameter. For a datum $x$, the function $f_\theta$ will produce a logit vector $z$. Given a loss function $\ell$, the target is to minimize the generalization loss which is $\mathbb{E}_{x\sim \mathcal{D}}\big[\ell(f_\theta(x), y \big]$, where $\mathcal{D}$ is the true distribution of the whole population. However, the parameter $\theta$ is often learnt from the sample $\mathcal{S}$ through the empirical loss function, which is $\sum_{i=1}^n \ell\big(f_\theta(x_i),y_i)\big)$.      


\subsection{Riemannian Geometry}

This section informally summarizes the key concepts of Riemannian Geometry that are used to describe the space of parameter $\theta$. We refer to Appendix \ref{appendix: riemannian manifolds} for a more rigorous discussion on Riemannian manifolds. Suppose the parameters $\theta$ belong to a \textit{low-dimensional manifold} $\mathcal{M}$ embedded in a higher-dimensional Euclidean ambient space $\mathcal{E}$. Denote $d = \text{dim } \mathcal{M}$, and $k = \text{dim } \mathcal{E}$. At any point $\theta$ on this manifold, there exists a \textbf{tangent space}, denoted by $T_\theta \mathcal{M}$, which can be thought of as a flat, linear space that "touches" the manifold at that point. By convention, the tangent space $T_\theta \mathcal{M}$ is an Euclidean subspace using the coordinate system with the current $\theta$ as an origin. For each $\theta$, we have an inner product $\left<\cdot, \cdot\right>_\theta$ defines on the tangent space $T_\theta \mathcal{M}$. When this metric varies smoothly by $\theta$, the manifold $\mathcal{M}$ is called the Riemannian manifold. 

In standard Euclidean space, the gradient of a function tells us the direction and rate of the steepest ascent direction. On a Riemannian manifold, we have an analogous concept called the \textbf{Riemannian gradient}, denoted by $\text{grad}_\theta f(\theta)$, which is an element of $T_\theta \mathcal{M}$ pointing in the direction of the steepest ascent but is adapted to the curvature of the manifold. This is essentially a gradient that respects the geometry of the manifold, ensuring that each step we take aligns with the manifold's shape and constraints.


When optimizing on a Riemannian manifold, we need to update parameters while staying "on" the manifold. A \textbf{retraction operation} at $\theta$, denoted by $\mathcal{R}_\theta(v)$, is an operation that moves from the tangent space back onto the manifold itself. Intuitively, the retraction operation can be thought of as an operation that gently places the current point back on the curved surface after a small step along the tangent space. Also, we often need to ensure that any updates or adjustments to our parameters remain within the structure of the manifold. This is where the \textbf{orthogonal projection} to the tangent space, denoted by $\mathcal{P}_0$, comes into play.  This operation projects any vector on the ambient space to the tangent space $T_\theta\mathcal{M}$, which is well-defined because $T_\theta \mathcal{M}$ is an Euclidean space. For short, we write $\mathcal{P}_\theta \mathbf{w}$, in which $\mathcal{P}_\theta$ is a symmetric matrix. Besides, we denote $\mathcal{B}^o_\theta(\rho;  T_\theta \mathcal{M})= \{\epsilon \in   T_\theta\mathcal{M} : \|\epsilon \|_2 \leq \rho\}$ a $\rho$-neighborhood in the tangent space at $\theta$, and $\mathcal{B}_\theta(\rho; \mathcal{M}) = \mathcal{R}_\theta(\mathcal{B}^o_\theta(\rho;  T_\theta \mathcal{M}))$ is the $\rho$-neighborhood on the manifold $\mathcal{M}$.

\subsection{Quotient manifolds}
We consider a collection of operations, that form a group $G$, acting on manifold $\mathcal{M}$. Formally, there exists a map from $G\times \mathcal{M}$ to $\mathcal{M}$ that satisfying the following properties: $(g_1g_2)\cdot\theta = g_1 \cdot (g_2\cdot \theta)$ and $e\cdot\theta =  \theta$, where $g_1,g_2$ and $e$ are elements of group $G$ with $e$ the identity element. For instance, let $\mathcal{M}$ be the space of matrices of rank $p$ in $\mathbb{R}^k$ and $G$ be the set orthogonal matrices in $\mathbb{R}^k$. The actions of $G$ on $\mathcal{M}$ form \textit{orbits} $[\theta] := \{g.\theta: g\in G \}$ in the space of $\theta$. The set of all possible orbits $[\theta]$, denoted by $\mathcal{M}/G$, formed the \textit{quotient manifold}. We have the following proposition (refer to \textbf{Theorem 21.10} in \citep{intromanifolds}):

\begin{proposition}
\label{theorem: quotient manifold}
    Suppose a Lie-group $G$ acts freely, smoothly, and properly on a smooth manifold $\mathcal{M}$. Then the orbit space $\mathcal{M}/G$ is a topological manifold of dimension equal to $\text{dim}(\mathcal{M}) - \text{dim}(G)$ with a unique smooth structure so that the quotient map $\pi: \mathcal{M} \to \mathcal{M}/G$ is a smooth submersion.
\end{proposition}



It is well known that when $\mathcal{M}$ is a Riemannian manifold, and $G$ is a Lie group acting on $\mathcal{M}$ freely and properly, then the quotient $\mathcal{M}/G$ is a manifold that inherits the Riemannian metric from $\mathcal{M}$ \citep{manifoldbook}. Given this property, we may denote $\mathcal{R}^Q_{[\theta]}, \mathcal{P}^Q_{[\theta]},$ and $\text{grad}^Q_{[\theta]}$ to be the retraction, projection, and the Riemannian gradient the quotient manifold, respectively. It is also worth noting that according to Proposition 4.1.3 \citep{matrixmanifolds}, the retraction operation, the projection, and the Riemannian gradient of the quotient manifold $\mathcal{M}/G$ are well-defined from the operations on the total space $\mathcal{M}$.

\subsection{Definitions of specific manifolds}

\textbf{Grassmann Manifold. \quad} The first manifold of interest in our study is the Grassmann manifold. To define it, we first consider the Stiefel manifold, which is given by:

\begin{definition}[The Stiefel Manifolds]
The set of $(n \times p)-$dimensional matrices, $p \leq n$, with orthogonal columns and Frobenius inner products forms a Riemannian manifold is called the \textit{Stiefel manifold} $St(p, n)$
\begin{equation*}
    St(p, n) \doteq \{\textbf{X} \in \mathbb{R}^{n \times p}: \textbf{X}^\top \textbf{X} = \textbf{I}_p\}.
\end{equation*}
\end{definition}

\cite{book} proposed multiple retractions for Stiefel manifolds. For the sake of computational complexity, we suggest using the retraction: $R_\textbf{X}(\varepsilon) = \text{qf}(\textbf{X} + \varepsilon)$ in which $\text{qf}(\textbf{A})$ denote the $\textbf{Q}$ factor of the decomposition of $\textbf{A} \in \mathbb{R}_*^{n \times p}$ as $\textbf{A} = \textbf{QR}$. The projection can also be derived as $\text{Proj}_\textbf{X}(\textbf{v}) = \textbf{v} - \textbf{X} \text{Sym}(\textbf{X}^\top \textbf{v})$ in which $\text{Sym}(\textbf{A}) = \frac{1}{2}(\textbf{A}+\textbf{A}^\top)$. 

In our applications, we are interested in $\mathcal{M} = St(n, p)$, and the Lie group $G$ is the orthogonal group $SO(p)$. A desirable property is that $SO(p)$ acts freely, properly on $St(n, p)$, then $Gr(n, p):= St(n, p)/SO(p)$ is a smooth quotient manifold by Proposition \ref{theorem: quotient manifold}, and is also Riemannian according to Page 233 by \cite{manifoldbook}. This quotient manifold is known as the \textit{Grassmann manifold}, which may also be defined as the set of all $p-$dimensional subspaces of an $n-$dimensional space. We refer to Appendix \ref{appendix: quotient manifolds} for more details about quotient manifolds and explicit formulas for operations on Grassmann manifolds that were implemented in the experiments. 

\textbf{SPD Manifold. \quad} Neural network constructions have been widely extended to the manifold of symmetric positive definite (SPD) matrices \cite{spdnet, rresnet}. Let SPD($n$) denote this manifold of $n \times n$ symmetric positive definite. Two standard metrics used for SPD(n) are the log-Euclidean metric \cite{log-euclid-spd}, which induces a flat structure on the matrices, and the canonical affine-invariant metric \cite{canonicalspd}, which induces non-constant negative sectional curvature. In our experiments, we are particularly interested in the log-Euclidean metric. We refer to Appendix \ref{appendix: spd manifold} for more details on the SPD manifold.

\section{Motivation}
\begin{figure}
    \centering
    \includegraphics[width=0.35\textwidth]{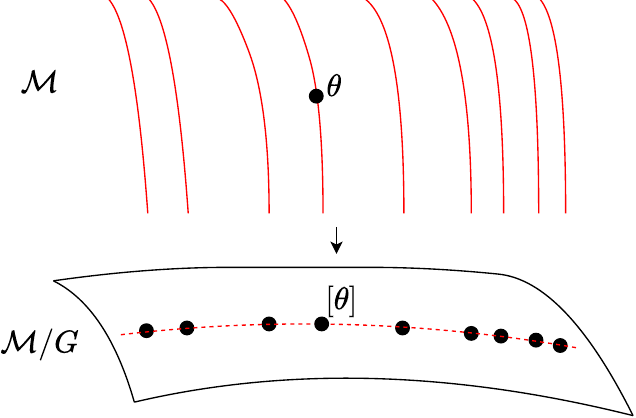}
    \caption{Example of quotient manifold reducing the dimensionality of the search space from 2D to the 1D curve.}
    \label{fig: quotient manifold}
    \vspace{-\baselineskip}
\end{figure}

\textbf{Exploiting group-invariance with quotient manifolds.\quad} In optimization, we often encounter scenarios where the loss function is invariant under the actions of a group $G$. Leveraging these group-invariant properties can lead to more efficient optimization strategies \cite{liu2023quotient}. For example, in 3D pose estimation, two objects may be considered identical up to a rotation; hence, optimizing over the entire 3D space to learn every pose is redundant. Instead, the problem can be simplified by quotient out the rotation and reducing the search space to a lower-dimensional manifold \citep{liu2023quotient}. Quotient manifold offers several advantages, such as dimensionality reduction, improved robustness, and enhanced generalization \citep{manifoldbook, Huang_2015_CVPR, siamesemanifold}. Figure \ref{fig: quotient manifold} provides a visual illustration of the quotient manifold. Recently, \cite{teleportoriginal, teleport2} developed the Teleportation technique to accelerate optimization. The core idea is to "teleport" to a new point in the parameter space at each iteration by applying the optimal group action that \textit{maximizes} the gradient norm, thus maximizing the update step size and speeding up convergence. However, a limitation of Teleportation is that it focuses on accelerating optimization rather than improving final performance, overlooking its impact on generalization as we will also show in Section \ref{section: experiments}, prior teleportation techniques \textit{does not} improve upon SGD in terms of the final performance. Subsequently, we will address this limitation with an \textit{opposite} teleportation mechanism designed to enhance generalization.

\textbf{Enhancing generalization ability.\quad} Sharpness-Aware minimization (SAM) \cite{sam} has shown superior performance across a wide range of applications to strengthen model generalization ability \citep{qu2022generalized, abbas2022sharp, chen2021vision, bahri-etal-2022-sharpness, cha2021swad}. The key idea of SAM is to minimize the sharpness term, which is defined as the worst-case loss value within a neighborhood of the current parameter. While standard SAM does not consider the intrinsic geometry of the parameter space, \cite{rsam} developed Riemannian-SAM that incorporates SAM for Riemannian manifolds. However, both SAM and Riemannian-SAM do not utilize the \textit{group-invariant properties} of the loss function, leaving room for improvement to enhance the generalization ability and model robustness further. In our method, we attempt to leverage this strength of the sharpness-aware methods to enhance generalization ability, and this benefit will be specifically highlighted in the experimental section \ref{section: domain adaptation} on Domain Adaptation.

\textbf{Combining everything.\quad} In our work, we propose to exploit the group-invariant structure of the loss function to strengthen the generalization ability and model robustness further. Let $l(\theta)$ denote the loss function; we assume that $\ell(\theta) = \ell(g \cdot \theta)$ for any group action $g \in G$. We also define $[\theta] = \{g\cdot\theta | g \in G\}$ as the \textbf{orbit} of $\theta$. To utilize this invariant property, each optimization iteration is decomposed into two phases: \textbf{1)} In the first phase; we \textit{move within the same orbit} to minimize the \textit{intra-orbit sharpness}. Specifically, we solve for the optimal group element $g^*$ that minimizes the sharpness at the point $g^*\cdot \theta$ where this sharpness is characterized by the gradient norm at $g \cdot(t\theta)$. This phase introduces an \textit{opposite} mechanism compared to Symmetry Teleportation \citep{teleportoriginal}. While Symmetry Teleportation aims to \textit{maximize the gradient norm} to accelerate optimization, it does not directly consider final performance. In contrast, our \textbf{sharpness-aware teleportation} mechanism focuses on \textit{minimizing the gradient norm} to improve model robustness and generalization. \textbf{2)} In the second phase, we \textit{move between different orbits} and minimize the \textit{inter-orbit sharpness}. Consider that the parameter space is a Riemannian manifold $\mathcal{M}$, the quotient space 
$\mathcal{M}/G$ is also a Riemannian manifold. By solving the geometry-aware sharpness-aware minimization problem on this quotient manifold, we jointly minimize sharpness across both phases, ensuring that both intra-orbit and inter-orbit sharpness are optimized. We hypothesize that decomposing each iteration into two steps and minimizing the sharpness at each step yields smaller sharpness than minimizing the sharpness across the full iteration. Indeed, this intuition will be supported by our ablation studies in Section \ref{ablation: sharpness} and theoretically justified by Theorem \ref{main theorem}.

\section{Methodology}
This section provides a theoretical analysis of our framework. Consider the optimization problem where the parameter space is a smooth manifold:
\begin{equation}
\label{eq: opt problem}
    \min_{\theta \in \mathcal{M}} L_\mathcal{D}(\theta).
\end{equation}
The condition $\theta \in \mathcal{M}$ can be interpreted as a constraint imposed on the parameters, for instance, orthogonality, full rank, or Euclidean. This work considers the case where the loss function is invariant under the actions of a Lie-group $G$:
\[\ell(f_\theta(x), y) = \ell(f_{g\cdot\theta}(x), y), \]
for all $g \in G, \theta \in \mathcal{M}$. Group-invariant loss functions appear in various applications, such as the Quotient Autoencoder for learning 3D shapes, with invariance under rotations \citep{liu2023quotient}, or affine-invariant loss functions for shape matching \citep{Simoncelli1995}. Under this invariance condition, the population loss $L_\mathcal{D}$ and the empirical loss $L_\mathcal{S}$ are also invariant under the actions of $G$, means that $L_\mathcal{D}(\theta) = L_\mathcal{D}(g\cdot\theta)$ and $L_\mathcal{S}(\theta) = L_\mathcal{S}(g\cdot\theta)$ for all $g \in G, \theta \in \mathcal{M}$.

Solving the optimization problem in Eq. (\ref{eq: opt problem}) encounters two challenges. Firstly, only a finite dataset $\mathcal{S}$ is available instead of the whole data distribution $\mathcal{D}$, therefore we do not have a direct evaluation of the general loss function $L_\mathcal{D}(\theta)$. Secondly, the parameters are constrained on Riemannian manifold $\mathcal{M}$ where the structure is typically more complicated than the Euclidean space. To tackle these problems, we present the following result showing the relation between the empirical and the population losses under the group-invariant loss property. The proof can be found in Appendix \ref{proof: main theorem}.   
\begin{theorem}
\label{main theorem}
     Under mild conditions, for any $\theta \in \mathcal{M}$, $\rho > 0$, and $\delta \in [0;1]$, with a probability of $1 - \delta$ over training set $\mathcal{S}$ generated from a distribution $\mathcal{D}$, it holds that:
   \begin{align*} \label{thm: main theorem}
    \displaystyle L_\mathcal{D}(\theta) &\leq \min_{g \in G}\max_{\theta' \in \mathcal{B}_{g\cdot\theta}(\rho; \mathcal{M})} L_\mathcal{S}(\theta') \\ &+  \mathcal{O}\Bigg(C_\mathcal{M}\rho^2 + \sqrt{\frac{ d + \log \frac{n}{\delta}}{n-1}}\Bigg),
\end{align*}
where $C_\mathcal{M}>0$ is a constant depending on the manifold $\mathcal{M}$.
\end{theorem}
This inequality expresses the generalization regarding neighborhood-wise training loss on the manifold instead of the whole ambient space as in \cite{sam}, and the Lie-group $G$. We note that $\mathcal{M}$ has a smaller intrinsic dimensionality than the ambient space dimension, $d \ll k$. Therefore, this theorem gives us a tighter bound of $\mathcal{O}(d^{1/2})$ compared to $\mathcal{O}(k^{1/2})$ of SAM \citep{sam}. Inspired by the inequality above, we propose to minimize the loss function: 
\begin{equation}
     L_\mathcal{S}^{RSAT}(\theta):= \min_{g\in G}\max_{\theta' \in \mathcal{B}_{g\cdot\theta}(\rho; \mathcal{M})}L_\mathcal{S}(\theta').
   \label{eq: loss rsat}
\end{equation}
This new loss function can be further unpacked as:

\begin{align*}
    L^{RSAT}_\mathcal{S}(\theta) = L_\mathcal{S}(\theta) + \mathfrak{S}(\theta;  G, \mathcal{M}),
\end{align*}
in which we have the sharpness term
\begin{equation*}
    \mathfrak{S}(\theta;  G, \mathcal{M}) = \min_{g \in G}\max_{\|\epsilon\|_2 \leq \rho, \epsilon\in  T_{g\cdot\theta}\mathcal{M}}  L_{\mathcal{S}}(\mathcal{R}_{g\cdot\theta}(\epsilon)) -  L_{\mathcal{S}}(\theta).
\end{equation*}
An important property of the second term is that it is invariant under group actions, as demonstrated by the following proposition, whose proof can be found in Appendix \ref{appendix: proof invariant sharpness}.
\begin{proposition}
\label{proposition: invariant sharpness}
    For any $g' \in G$, we have
    \[\mathfrak{S}(\theta; G, \mathcal{M}) = \mathfrak{S}(g'\cdot\theta; G, \mathcal{M}).\]
\end{proposition}

Motivated by this proposition, the term $\mathfrak{S}(\theta;G, \mathcal{M})$ will be referred to as the \textbf{group-invariant sharpness}. Thus, to minimize this loss function, we propose to take two gradient steps SAM-like update: one \textit{ascend-teleportation step} which solves for $g$ and $\theta'$ that minimizes the group-invariant sharpness and updates $\theta' = \mathcal{R}_{g\cdot\theta}(\epsilon)$, followed by a \textit{Riemannian descending step} that solves Eq. (\ref{eq: loss rsat}) with the current value of $\theta'$. To find the explicit form for $g$ and $\epsilon$, the sharpness term can be unpacked as:
\begin{align}
     \mathfrak{S}(\theta; G, \mathcal{M})
    \label{eq: minimax group-invariant sharpness}
    &= \min_{g \in G}\max_{\|\epsilon\| \leq \rho, \epsilon \in T_{g\cdot\theta}\mathcal{M}}\left<\text{grad}_{g\cdot\theta} L_{\mathcal{S}}(g\cdot\theta), \epsilon \right>_{g\cdot\theta}.
\end{align}

We refer to Appendix \ref{appendix: group-invariant sharpness derivation} for more details of the derivation. The Riemannian metric can be expressed as $\left <\epsilon, \epsilon' \right >_\theta = \epsilon^\top \textbf{D}_\theta \epsilon'$ for some matrix $\mathbf{D}_\theta$ that reflects the local geometry at $\theta$. The closed-form solution to the inner maximization problem can be found in Appendix \ref{appendix proof: closed from max}. However, computing this solution is prohibitively expensive. Instead, we propose a more practical approach that first solves:

\begin{equation}
\label{eq: relaxed problem}
    \overline{\epsilon} = \arg\max_{\|\epsilon\|\leq\rho}\text{grad}_{g\cdot\theta}L_\mathcal{S}(g\cdot\theta)^\top \mathbf{D}_{g\cdot\theta} \epsilon.
\end{equation}

The given solution will then be projected onto the tangent space to get $\epsilon^* = \mathcal{P}_{g\cdot\theta}\overline{\epsilon}$. Eq. (\ref{eq: relaxed problem}) yields the following solution, whose proof can be found in Appendix \ref{appendix proof: related solution}.

\begin{proposition}\label{alg: relaxed}
    The solution to the maximization problem in Eq. (\ref{eq: relaxed problem}) is given by:
    \begin{equation*}\label{prep: approximated}
    \overline{\epsilon} = \rho
    \frac{\mathrm{grad}_{g\cdot\theta}(L_\mathcal{S}(g\cdot\theta))^\top \mathbf{D}_{g\cdot\theta}}
    {\left \|  \mathrm{grad}_{g\cdot\theta}(L_\mathcal{S}(g\cdot\theta))^\top \mathbf{D}_{g\cdot\theta} \right \|
    }.
    \vspace{-\baselineskip}
    \end{equation*}
\end{proposition}
With the value $\bar{\epsilon}$ above, the solution to the outer minimization problem in Eq. (\ref{eq: minimax group-invariant sharpness}) becomes
\begin{align*}
    g^* &= \arg\min_{g\in G} \|\text{grad}_{g\cdot\theta}L_\mathcal{S}(g\cdot\theta)^\top\mathbf{D}_{g\cdot\theta}\| \\
    &= \arg\min_{g\in G} \|(\mathcal{P}_{g\cdot\theta}\nabla L_\mathcal{S}(g\cdot\theta))^\top \mathbf{D}_{g\cdot\theta}\|,
\end{align*}
which can be solved efficiently with SGD. Once we got $g^*$ and $\epsilon^*$ for the ascending step, plugging in $\theta' = \mathcal{R}_{g^*\cdot\theta}(\epsilon^*)$ to Eq. (\ref{eq: loss rsat}), we perform the descending step that solves for:
\begin{equation*}
    \theta^* = \arg\min_{\theta \in \mathcal{M}} L_\mathcal{S}(\mathcal{R}_{g^*\cdot\theta}(\epsilon^*)),
\end{equation*}
which can be done with Riemannian SGD \citep{rsgd}. However, we recall the condition that the loss function is invariant under the group action. This condition suggests quotient out the optimization problem into a smaller space. Instead of performing the descending step on the whole space, we propose moving on the quotient manifold to move \textit{between orbits}. To do so, consider another "view" of the loss function on the quotient manifold, which we define the \textit{quotient loss} as:
\begin{equation*}
    L^Q_\mathcal{S}([\theta]) := L_\mathcal{S}(\theta')
\end{equation*}
for all $\theta' \in [\theta]$. Under the condition that $L_\mathcal{S}$ is invariant under the $G-$actions, it implies that $L^Q_\mathcal{S}$ is well-defined with the domain being the quotient manifold. Recall that when $G$ acts freely, properly on $\mathcal{M}$, the quotient manifold is also Riemannian. Thus, we can perform the descent step on the quotient loss $L^Q_\mathcal{S}([\mathcal{R}_{g^*\cdot\theta}(\epsilon^*)])$. The main advantage of employing the quotient manifold is that it enables a more efficient optimization process by "compressing" the search space to a lower dimensional space, moving naturally between the orbits instead of point-to-point. The idea of optimizing on the quotient manifold has also been previously employed by \cite{siamesemanifold} and has demonstrated improved efficiency and generalization. For more details on the concrete advantages of quotient out the optimization problem, we refer to the Appendix \ref{appendix: advantages}. Motivated by the theoretical development above, we propose the Riemannian Sharpness-Aware Teleportation (RSAT) given in Alg. \ref{alg: rsat}, which simultaneously minimizes the loss function and the group-invariant sharpness while restricting the search space within the quotient manifold. For more clarity, Figure \ref{fig: schematic} illustrates the schematic of parameters update of RSAT compared to SGD. 

\textbf{Comparison to Symmetry Teleportation \cite{teleportoriginal}. \quad} It is noteworthy that our teleportation mechanism is \textit{opposite} to the Symmetry Teleportation proposed by \cite{teleportoriginal}. This difference originates from the different purposes of teleporting in our algorithm. In particular, Symmetry Teleportation seeks for the group action that \textit{maximizes} the gradient norm to accelerate optimization but does not take into account the final performance and the generalization ability. On the other hand, RSAT finds the group action that \textit{minimizes} the gradient norm with the purpose of \textit{improving generalization ability}, thereby enhancing the final performance.

\begin{algorithm}[H]
\caption{Riemannian Sharpness-aware Teleportation (RSAT)}
\label{alg: rsat}
\begin{algorithmic}
\State {\bfseries Input:} Parameter space $\mathcal{M}$, Lie group $G$ acting on $\mathcal{M}$, training set $\mathcal{S} \doteq \cup^n_{i=1} \{(x_i, y_i) \}$, batch size $b$, learning rate $\eta > 0$, ascent step sizes $\lambda > 0$. 
\State Initialize $\theta_0 \in \mathcal{M}$, $t = 0$
   \Repeat
    \State Sample mini batch $\mathcal{B} = \{(x_i, y_i)\}_{i=1}^b$
    \State $g^* = \arg\min_{g \in G} \|(\mathcal{P}_{g\cdot\theta_t}\nabla L_\mathcal{B}(g\cdot \theta_t))^\top \mathbf{D}_{g\cdot\theta_t}\|$
    \State \textit{Teleport step:} $\overline{\theta}_t = g^*\cdot\theta_t$
    \State $\overline{\epsilon} = \lambda\frac{(\mathcal{P}_{ \overline{\theta}_t}\nabla L_\mathcal{S}( \overline{\theta}_t))^\top\mathbf{D}_{ \overline{\theta}_t}}{\|(\mathcal{P}_{ \overline{\theta}_t}\nabla L_\mathcal{S}( \overline{\theta}_t))^\top\mathbf{D}_{ \overline{\theta}_t}\|}$
    \State $\epsilon^* = \mathcal{P}_{\overline{\theta}_t}(\overline{\epsilon})$
    \State \textit{Ascend-teleport step:} $\hat{\theta}_t = \mathcal{R}_{ \overline{\theta}_t}(\epsilon^*)$ 
    \State \textit{Descend step:} \\$[\theta'_{t+1}] = \mathcal{R}^Q_{\theta_t}\big(-\eta \text{grad}^Q_{\theta_t} \big( L^Q_{\mathcal{B}}([\hat{\theta}_t])\big)\big)$, and pick representative $\theta_{t+1} \in [\theta'_{t+1}]$
   \Until{\textit{converges}}
\end{algorithmic}
\end{algorithm}

\section{Experiments}
\label{section: experiments}
\subsection{Image Classification} 
\label{metric learning}
In this first experiment, we evaluate RSAT against other baselines on a standard image classification task. It is important to note that, in this setting, the parameter space of the entire network is not necessarily Riemannian. However, the following sections will demonstrate the practical advantages of Riemannian manifolds when imposing certain conditions on specific subsets of parameters, as well as the effectiveness of our method in optimizing Riemannian parameters.

\textbf{Metric Learning with Grassmann Manifolds. \quad} We consider two self-supervised settings, including \textbf{labeled self-supervised learning} with Supervised Contrastive (SupCon) methodology proposed by \cite{supcon} and \textbf{unlabeled self-supervised learning} with SimCLR loss function \citep{simclr}. For a set of $N$ randomly sampled sample/label pairs, $\{\mathbf{x}_k, \mathbf{y}_k \}_{k = 1}^{N}$, the corresponding batch used for training consists of $2N$ pairs, $\{\Tilde{\mathbf{x}}_l, \Tilde{\mathbf{y}}_l \}_{l =1}^{2N}$, where $\Tilde{\mathbf{x}}_{2k}$ and $\Tilde{\mathbf{x}}_{2k-1}$ are random augmentations of $\mathbf{x}_k$, and $\Tilde{\mathbf{y}}_{2k-1} = \Tilde{\mathbf{y}}_{2k} = \mathbf{y}_k$. A set of $N$ samples is referred to as a "batch," and the set of $2N$ samples is a "multiview batch." Within a multiview batch, let $i \in I = \{1, \cdots, 2N\}$ be the index of an arbitrary augmented sample, and let $j(i)$ be the index of the other augmented sample originating from the same source sample. The architecture of both settings involves two components: \textbf{1)} The backbone Encoders, which is denoted as $\text{Enc}(\cdot)$; and \textbf{2)} The projection head $P(\cdot)$, which is either a linear or fully-connected low-dimensional layer. It is worth noting that the projection head $P(\cdot)$ differs from the Riemannian projection operation $\mathcal{P}_\mathbb{\theta}$. For any $l$, denote $\mathbf{z}_l = P(\text{Enc}(\Tilde{\mathbf{x}}_l))$.

\label{section: applications}

\begin{figure}
    \centering
        \includegraphics[width=0.48\textwidth]{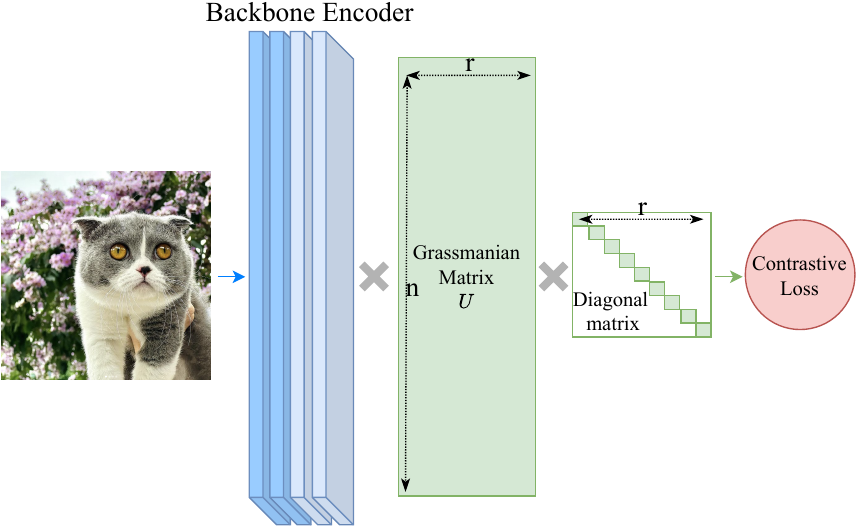}

  \caption{Metric learning where the typical projectional layer is replaced with the multiplication to $\mathbf{U} \in \text{Gr}(n, p)$ and $\mathbf{S}$ diagonal.}      \label{fig: architecture}
\vspace{-\baselineskip}
\end{figure}

In labeled self-supervised learning, the logit are trained with the SupCon objective proposed by \cite{supcon}: 
    \begin{equation*}
        \mathcal{L}^{sup}_{out}:= \sum_{i \in I}\frac{-1}{|C(i)|} \sum_{p \in C
        (i)}\log \frac{\exp(\frac{\mathbf{z}_i \cdot \mathbf{z}_p}{\tau})}{\sum_{a \in A(i)}\exp(\frac{\mathbf{z}_i \cdot \mathbf{z}_a}{\tau})}
    \end{equation*}
    \begin{equation*}
        = \mathcal{L}\big(\mathbf{z}_1 \cdots, \mathbf{z}_{2N}) = \mathcal{L}(P(f(\Tilde{\mathbf{x}}_1)) \cdots, P(f(\Tilde{\mathbf{x}}_{2N}))\big), 
    \end{equation*}
    where $A(i) = I \backslash  \{i\}$, and $C(i) = \{c \in A(i): \Tilde{\mathbf{y}}_c = \Tilde{\mathbf{y}}_i\}$. 
    
On the other hand, SimCLR \citep{simclr} defines the loss for a positive pair of examples as: 
\begin{equation*}
    \ell_{i, j} = -\log\frac{\exp(s_{i, j}/\tau)}{\sum^{2N}_{k = 1} \mathbbm{1}_{k \neq i} \exp(s_{i, k}/\tau)},
\end{equation*}
 where $s_{i, j} = \frac{\mathbf{z}_i \cdot \mathbf{z}_j}{\|\mathbf{z}_i\| \|\mathbf{z}_j\|}$ measures the similarity of the two logits, and $\mathbbm{1}_{[k \neq i]}$ is an indicator function evaluating to $1$ iff $k \neq i$. Then, the final loss is computed across all positive pairs in a mini-batch.
 
    In our applications, the Euclidean inner product is replaced with the Mahalanobis distance $\left<\mathbf{h}_1, \mathbf{h}_2 \right> = \mathbf{h}^\top_1\mathbf{M}\mathbf{h}_2$, in which $\mathbf{M}$ is learnable to take into account the local geometry. Singular Value Decomposition yields $\mathbf{M} = \mathbf{U}\mathbf{D}\mathbf{U}^{\top} = \mathbf{U}\mathbf{D}^{1/2}\mathbf{D}^{1/2}\mathbf{U}^\top$. Denote $\mathbf{S} = \mathbf{D}^{1/2}$, it follows that:
    \begin{equation*}
        \left<\mathbf{h}_1, \mathbf{h}_2 \right> = \mathbf{h}^{\top}_1\mathbf{M}\mathbf{h}_2 = (\mathbf{h}_1^{\top}\mathbf{U}\mathbf{S} ) \cdot (\mathbf{h}_2^{\top}\mathbf{U}\mathbf{S})^{\top}.
    \end{equation*}
    Motivated by the equation above, instead of optimizing $\mathcal{L}(P(\text{Enc}(\Tilde{\mathbf{x}}_1)), \cdots, P(\text{Enc}(\Tilde{\mathbf{x}}_{2N})))$, we will optimize $\mathcal{L}(P(\text{Enc}(\Tilde{\mathbf{x}}_1))\mathbf{US}, \cdots, P(\text{Enc}(\Tilde{\mathbf{x}}_{2N}))\mathbf{US})$ in which $\mathbf{U}$ is a rotational matrix on the Stiefel manifold, and $\mathbf{S}$ is a diagonal matrix. Notice that if we have $\textbf{K} \in SO(p)$, then $\textbf{UK} \in St(n, p)$. 
    Moreover, the SimCLR and the SupCon loss functions are \textit{invariant under the right actions of $SO(p)$ on the orthogonal matrix $\textbf{U}$}, which we refer to Appendix \ref{proof: invariant} for the proof. So, the search space of $\textbf{U}$ is reduced from the Stiefel manifold to the lower-dimensional Grassman manifold. Since $\mathbf{U}$ lies on the Grassmann manifold, this matrix will be trained with RSAT.  Other parameters, including the backbone and the diagonal matrix $\mathbf{S}$, will be trained via Euclidean optimizers such as SAM or SGD. In our implementation, to solve for $\mathbf{K}$, we adopt the Cayley transformation $\mathbf{K} = (\mathbf{I} + \mathbf{Q})(\mathbf{I} - \mathbf{Q})^{-1}$ where $\mathbf{Q}$ is a skew-symmetric matrix satisfying $\mathbf{Q} = -\mathbf{Q}^\top$ and is initialized to $\mathbf{0}$. The new architecture is illustrated in Figure \ref{fig: architecture}.

\textbf{Experimental Results. \quad}
We experimented with four vision datasets, including CIFAR10, CIFAR100, STL10, and Aircraft. The performance of RSAT is compared with Euclidean optimization methods, including SGD, SAM \citep{sam}, Symmetry Teleportation \citep{teleportoriginal}; Riemannian methods, including RSGD \citep{rsgd}, Riemannian-SAM \citep{rsam}; and the orthogonal manifold optimizer OPT \cite{opt}. The models are trained with a batch size of $256$ on CIFAR100, CIFAR10, and STL10 and $64$ on FGVCAircraft. Throughout the experiments, $\mathbf{U}$ was chosen to have the size $d \times 32$, where $d$ depends on the output size of the backbone. We conducted five independent runs for each setting to measure the error and reported the means along with the 95\% confidence interval. For more experimental details, please refer to Appendix \ref{appendix: experimental details}. 

 As discussed in previous sections, we examine the classification accuracy of labeled and unlabeled self-supervised learning settings. The experiment on labeled self-supervised learning has two stages. The model is trained with the SupCon loss in the pretraining stage using the baselines and RSAT. The second stage freezes the parameters and trains a linear classifier, i.e., linear evaluation. In the pre-trained step, the Riemannian optimizers, including RSAT, Riemannian-SAM, and RSGD, are applied to train the Grassmannian parameters, which is the matrix \textbf{U} as discussed in Section \ref{section: applications}. We especially note that RSAT made a notable improvement upon SupCon with SGD by more than 8\% on STL10 and 6\% on CIFAR100, which is shown in Table \ref{tab: supcon}. Similarly, the experiment on unlabeled self-supervised learning also has two stages. On average, RSAT outperforms the baselines by a notable margin. For instance, RSAT improves upon SGD by a margin of 7\%, as shown in Table \ref{tab: simclr}. It is also noteworthy that in both experiments, Symmetry Teleportation \textit{does not} improve the final performance upon vanilla SGD and introduce larger deviation, therefore highlighting the benefits of our new teleportation mechanism in enhancing generalization ability.

\begin{table}[h]
\centering
\resizebox{\columnwidth}{!}{%
\begin{tabular}{llcccc}
\midrule
Model     &   Method  & CIFAR100 & CIFAR10 & STL10 & Aircraft \\ \midrule \midrule

  &   SGD     &    $74.08_{\pm .21}$        &  $95.96_{\pm .19}$  & $85.69_{\pm .23}$ &  $78.19_{\pm .33}$   \\
            &   Teleport     &    $76.10_{\pm .40}$        &  $94.99_{\pm .25}$  & $86.23_{\pm .44}$ &  $78.02_{\pm .48}$   \\
            &   SAM     &    $76.93_{\pm .18}$      & $96.07_{\pm .22}$ & $87.10_{\pm .21}$ & $81.73_{\pm .24}$   \\
 ResNet34                 &   OPT    &    $\underline{78.51}_{\pm .86}$      & $94.01_{\pm .17}$ & $\underline{89.10}_{\pm .37}$ & $82.24_{\pm .41}$   \\
            &   \multicolumn{1}{l}{RSGD}  & $77.32_{\pm .24}$ & $\underline{96.25}_{\pm .10}$ & $86.03_{\pm .18}$& $83.17_{\pm .35}$   \\
            &   \multicolumn{1}{l}{R-SAM} & $78.43_{\pm .19}$  &  $95.80_{\pm .21}$ & $87.21_{\pm .23}$ & $\underline{84.37}_{\pm .15}$  \\ 
            &   \multicolumn{1}{l}{\textbf{RSAT}}  & $\textbf{80.35}_{\pm .31}$   & $\textbf{96.64}_{\pm .19}$  & $\textbf{89.13}_{\pm .20}$ & $\textbf{87.52}_{\pm .20}$ \\ \midrule
ResNet50   &   SGD   &  $75.29_{\pm .21}$ & $95.99_{\pm .09}$ & $83.33_{\pm .22}$& $82.03_{\pm .21}$    \\ 
&   Teleport   &  $74.84_{\pm .53}$ & $96.21_{\pm .04}$ & $82.78_{\pm .34}$& $83.32_{\pm .40}$    \\ 
            &   SAM   &  $76.73_{\pm .18}$ & $\underline{96.31}_{\pm .11}$ & $85.02_{\pm .19}$ & $82.84_{\pm .24}$  \\ 
            &   OPT   &  $78.32_{\pm .81}$ & $93.17_{\pm .42}$ & $82.10_{\pm .96}$ & $81.13_{\pm .41}$  \\ 
            &   \multicolumn{1}{l}{RSGD}  &  $78.13_{\pm .24}$  & $96.06_{\pm .18}$ & $84.23_{\pm .15}$ & $83.38_{\pm .31}$   \\
            &   \multicolumn{1}{l}{R-SAM}  &  $\underline{79.46}_{\pm .12}$  & $96.11_{\pm .23}$ & $\underline{87.35}_{\pm .20}$ & $\underline{84.02}_{\pm .15}$   \\
            &   \multicolumn{1}{l}{\textbf{RSAT}}  & $\textbf{81.52}_{\pm .19}$   & $\textbf{96.88}_{\pm .13}$ & $\textbf{91.72}_{\pm .12}$ & $\textbf{85.52}_{\pm .31}$  \\ \bottomrule
\end{tabular}
}
\caption{Classification accuracy on labeled self-supervised setting with SupCon loss.}
\label{tab: supcon}
\vspace{-\baselineskip}
\end{table}

\begin{table}[h]
\centering
\resizebox{\columnwidth}{!}{%
\begin{tabular}{llcccc}
Model     &   Method  & CIFAR100 & CIFAR10 & STL10 & Aircraft \\ \midrule \midrule

ResNet34    &   SGD     &   $63.05_{\pm .31}$        &  $90.98_{\pm .22}$  & $59.37_{\pm .43}$ &  $64.23_{\pm .30}$   \\
&   Teleport     &   $64.27_{\pm .52}$        &  $89.73_{\pm .48}$  & $58.72_{\pm .69}$ &  $65.23_{\pm .64}$   \\
            &   SAM     & $64.32_{\pm .31}$  & $91.16_{\pm .20}$ & $67.83_{\pm .36}$ & $\underline{64.01}_{\pm .39}$  \\

            &   OPT     & $\underline{69.12}_{\pm .13}$  & $\underline{92.87}_{\pm .06}$ & $68.13_{\pm .64}$ & $61.22_{\pm .13}$  \\
            
            &   \multicolumn{1}{l}{RSGD } &  $63.58_{\pm .22}$ & $91.02_{\pm .30}$ & $67.02_{\pm .44}$ & $63.93_{\pm .22}$  \\ 
            &   \multicolumn{1}{l}{R-SAM } & $67.26_{\pm .27}$  & $92.14_{\pm .29}$ & $\underline{70.08}_{\pm .28}$ & $61.82_{\pm .34}$  \\ 
            &   \multicolumn{1}{l}{\textbf{RSAT}}  & $\textbf{71.04}_{\pm .24}$ & $\textbf{93.47}_{\pm .24}$  & $\textbf{73.81}_{\pm .29}$ & $\textbf{65.66}_{\pm .41}$   \\ \midrule
ResNet50   &   SGD   & $65.65_{\pm .33}$ & $92.98_{\pm .27}$ & $65.35_{\pm .18}$ & $61.20_{\pm .41}$   \\ 
&   Teleport   & $65.02_{\pm .71}$ & $93.24_{\pm .41}$ & $65.37_{\pm .29}$ & $61.94_{\pm .66}$   \\ 
            &   SAM   & $67.24_{\pm .19}$ & $93.11_{\pm .25}$ & $69.93_{\pm .20}$ & $63.17_{\pm .44}$    \\ 
            &   OPT   & $68.33_{\pm .02}$ & $89.96_{\pm .17}$ & $68.34_{\pm .03}$ & $\underline{66.29}_{\pm .44}$    \\ 
            &   \multicolumn{1}{l}{RSGD } & $66.31_{\pm .33}$  & $93.50_{\pm .30}$ & $68.61_{\pm.39}$ & $65.01_{\pm .44}$ \\ 
            &   \multicolumn{1}{l}{R-SAM}  & $\underline{69.12}_{\pm .35}$ & $\underline{94.27}_{\pm .28}$ & $\underline{70.39}_{\pm .15}$ & $64.97_{\pm .19}$   \\
            &   \multicolumn{1}{l}{\textbf{RSAT}}  & $\textbf{73.20}_{\pm .31}$ & $\textbf{96.04}_{\pm .35}$ & $\textbf{72.54}_{\pm .29}$ & $\textbf{68.34}_{\pm .33}$  \\ \bottomrule
\end{tabular}
}
\caption{Classification accuracy on unlabeled self-supervised setting with SimCLR loss.}
\label{tab: simclr}
\vspace{-\baselineskip}
\end{table}

We also note that RSAT is expected to be slower than SAM due to the additional Riemannian computation. However, the difference is negligible, as shown in Appendix \ref{ablation: runtime}. Compared to Riemannian-SAM, even though RSAT involves the additional teleportation step, it is compensated by a more efficient optimization over the quotient manifold. Moreover, in practice, only a few teleportation steps are sufficient for a notable performance improvement (i.e., five descent steps as implemented in our experiments). Therefore, the runtime is practically around the same as Riemannian-SAM, and RSAT incurs no additional computational overhead compared to SAM and R-SAM, maintaining the same theoretical complexity and scalability.

\subsection{Domain Adaptation.}

\label{section: domain adaptation}


As discussed in previous sections, both steps in each iteration of our method jointly aim to improve generalization by minimizing sharpness. To strengthen our claim that RSAT enhances generalization ability, we evaluate RSAT in cross-dataset and cross-domain settings using the Grassmann manifold. Similar to the previous section, we integrate the Grassmann manifold with ResNet34 for metric learning. In this experiment, we compare RSAT against Riemannian-SAM and SAM. For the cross-dataset setting, we consider CIFAR10 (\textbf{C10}), CIFAR100 (\textbf{C100}), and STL10 (\textbf{S10}), which are natural scene datasets. For the cross-domain setting, we evaluate the methods on the Office-31 dataset, which consists of three domains: Amazon (\textbf{A}), Webcam (\textbf{W}), and DSLR (\textbf{D}). As shown in Table \ref{tab:cross-dataset}, RSAT consistently outperforms both Riemannian-SAM and SAM across all dataset pairs, underscoring the effectiveness of our two-step iterative sharpness-aware approach in improving generalization.

\begin{table}[h]
\centering
\vspace{-4mm}
\resizebox{\linewidth}{!}{%
\begin{tabular}{c|ccc|cccccc}
\hline
     & \multicolumn{3}{c|}{Cross-dataset} & \multicolumn{6}{c}{Cross-domain (Office-31)} \\ \hline
 &
  C10 -\textgreater S10 &
  S10 -\textgreater C10 &
  S10 -\textgreater C100 &
  W -\textgreater D &
  W -\textgreater A &
  D -\textgreater A &
  D -\textgreater W &
  A -\textgreater D &
  A -\textgreater W \\ \hline
SAM  & 81.2       & 61.0      & 29.0      & 36.7  & 22.8 & 32.54 & 41.29 & 26.21 & 27.39 \\
RSAM & 82.5       & 61.5      & 29.1      & 38.9  & 26.4 & 34.74 & 43.00 & 26.43 & 28.05 \\ \hline
RSAT &
  \textbf{83.2} &
  \textbf{61.9} &
  \textbf{29.4} &
  \textbf{42.3} &
  \textbf{26.6} &
  \textbf{35.86} &
  \textbf{47.28} &
  \textbf{28.43} &
  \textbf{28.55} \\ \hline
\end{tabular}
}
\caption{Cross-dataset and cross-domain performance on the Natural Scene datasets and Office-31 dataset.}
\label{tab:cross-dataset}
\vspace*{-\baselineskip}
\end{table}

\subsection{Video Classification with the SPD manifold}
After demonstrating the effectiveness of RSAT for image classification using the Stiefel manifold, we extend our experiments to the fully Riemannian network setting on four video classification datasets: AFEW \citep{afew}, FPHA \citep{FPHA}, NTU RGB+D \citep{ntu}, and HDM05 \citep{hdm05}. A common application of SPD manifold-based models is learning over full-rank covariance matrices, which reside on the manifold of SPD matrices. In this experiment, we leverage the SPD manifold to optimize the Riemannian Residual Recurrent Network (RResNet) \cite{rresnet} using the Log-Euclidean metric. Notably, if $M$ is an SPD matrix and $K \in SO(p)$ is an orthogonal matrix, then $MK$ remains an SPD matrix. In other words, $SO(p)$ acts on SPD($n$). Thus, as in the previous experiment, we may teleport on $G = SO(p)$. Since this task involves optimizing a fully Riemannian network, we compare RSAT against Riemannian optimizers, including Riemannian-SAM and RSGD. Table \ref{tab: spd} presents the results, showing that RSAT consistently outperforms the baselines, with notable performance gains in certain cases—such as a 4\% improvement on FPHA compared to Riemannian-SAM.

\begin{table}[ht]
\centering
\resizebox{0.8\linewidth}{!}{%
\begin{tabular}{c|cccc}
\hline
Method & AFEW           & FPHA           & NTU RGB+D      & HDM05          \\ \hline
R-SGD & 34.20          & 61.13          & 40.07          & 69.16          \\
R-Adam & 36.38          & 64.58          & 42.99          & 69.80          \\
R-SAM  & 37.22          & 62.19          & 41.78          & 70.01          \\
RSAT   & \textbf{38.23} & \textbf{66.99} & \textbf{43.15} & \textbf{72.64} \\ \hline
\end{tabular}
}
\caption{Performance with SPD manifold on video classification datasets. The performance of R-Adam is directly quoted from [1].}
\label{tab: spd}
\vspace{-\baselineskip}
\end{table}

\section{Ablation Studies}
In this section, we study the effectiveness of our RSAT in minimizing model sharpness and enhancing model robustness to parameter perturbation.

\begin{figure}[ht]
    \centering
    \includegraphics[width = 0.5\textwidth]{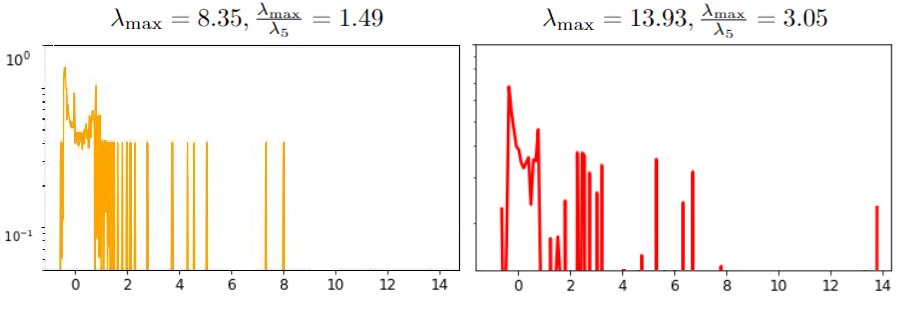}
    \caption{Hessian spectra of RSAT {\color{YellowOrange}(yellow)} vs. Riemannian-SAM {\color{red}(red)}. For RSAT, $\lambda_{\max} = 8.35, \frac{\lambda_{\max}}{\lambda_5} = 1.49$. For Riemannian-SAM, $\lambda_{\max} = 13.93, \frac{\lambda_{\max}}{\lambda_5} = 3.05$.}
    \label{fig: Hessian spectral}
    \vspace{-\baselineskip}
\end{figure}

\textbf{Hessian Spectra. \quad}Throughout the paper, we have designed RSAT to seek a region with low sharpness in the context of Riemannian manifolds. In this section, to verify the capability of RSAT to find the low-sharpness region, we contrast the spectrum of the Hessian of RSAT with Riemannian-SAM, which is also a flat optimizer on the Riemannian manifolds. The methods were trained on CIFAR100 with ResNet34 for 400 epochs. Figure \ref{fig: Hessian spectral} shows that the model trained with RSAT establishes a notably lower maximum eigenvalue (8.35 of RSAT vs. 13.93 of Riemannian-SAM). We also measure the bulk of the spectrum (the ratio $\lambda_{\max}/\lambda_5$, a metric commonly used as a proxy for sharpness in prior works \citep{sharpnessbulk, sam}), which are up to $3.05$ for Riemannian-SAM and is only $1.49$ for RSAT, which altogether suggest that RSAT converges to minima having lower curvature. 
\label{ablation: sharpness}

\begin{figure}[ht]
    \centering
    \includegraphics[width=0.35\textwidth]{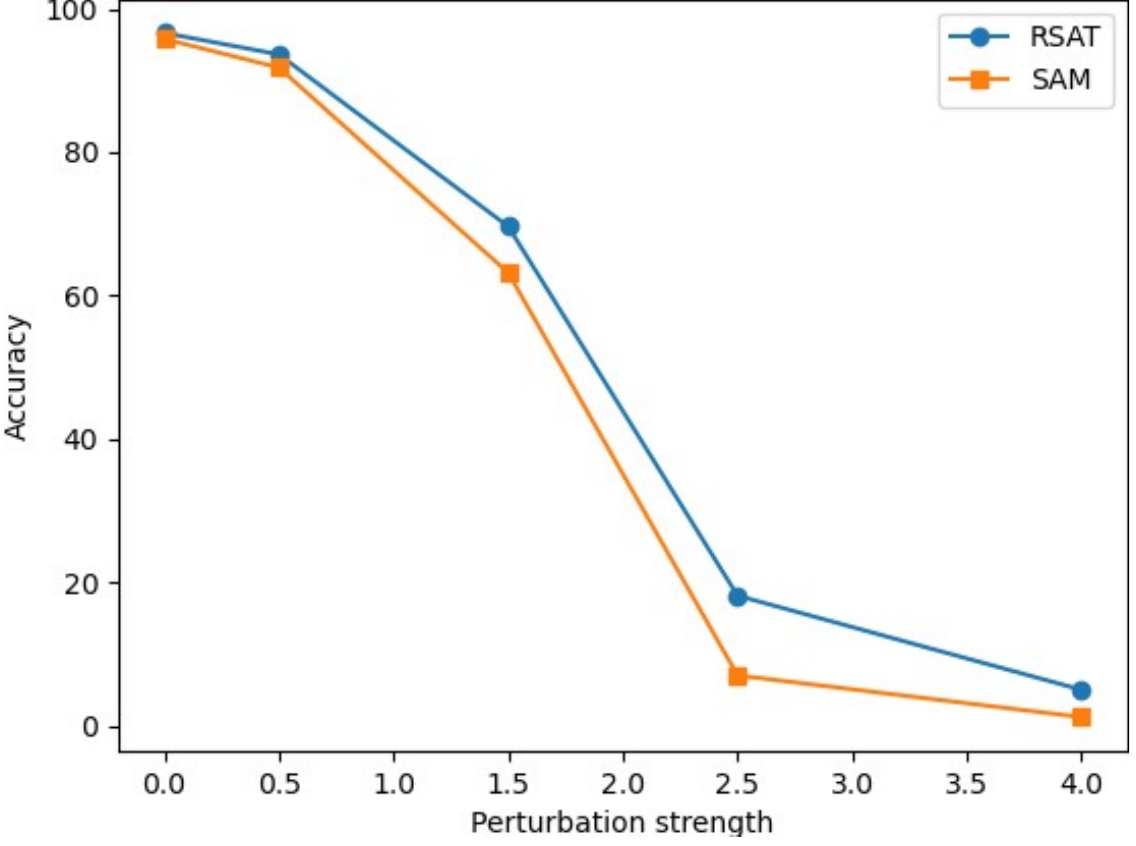}
    \caption{Robustness with Adversarial parameter perturbation}
    \vspace{-\baselineskip}
    \label{fig: robustness}
    \vspace{-\baselineskip}
\end{figure}

\textbf{Model Robustness to Parameter Perturbation. \quad} Besides the generalization ability, another important property of the proposed approach is the robustness of the trained model. Recently, to measure the vulnerability of neural networks, the adversarial perturbation (Sun et al., 2021) was proposed, where they consider the worst-case scenario under parameter corruption, which amounts to perturbation along the gradient direction. In particular, the perturbation was given by $\displaystyle\theta' = \theta + \alpha \frac{\nabla L_\mathcal{S}(\theta)}{\|\nabla L_\mathcal{S}(\theta)\|}$. In this section, we apply this perturbation to ResNet50 trained by RSAT and SAM on CIFAR10, with the perturbation strength $\alpha$ varying from $0$ to $4.0$. According to Figure \ref{fig: robustness}, RSAT demonstrates less performance degradation as $\alpha$ varies compared to SAM.


\section{Conclusion}
We have expanded the concept of flat minimizers to differential manifolds with a novel Riemannian optimizer that integrates teleportation, quotient manifolds, and sharpness-aware minimization in the Riemannian context. This algorithm is supported by a theorem that characterizes generalization ability by bounding the population loss with the worst-case empirical loss within a neighborhood on the manifold. RSAT has proven its effectiveness across various settings, consistently outperforming comparative methods.

\bibliographystyle{plainnat}
\bibliography{rsat}

\newpage
\clearpage
\appendix
\setcounter{page}{1}
\maketitlesupplementary

\section{Wall-clock runtime}
\label{ablation: runtime}
This section compares the single-epoch wall-clock runtimes of SGD, SAM, Riemannian-SAM, RSGD, and RSAT. SAM, Riemannian-SAM, and RSAT are expected to take at least twice as long as SGD or RSGD because these methods involve double backward-forward computation in each iteration. We note that even though RSAT involves additional computation for the teleportation step, it is compensated by reducing the search space to the quotient manifold. Hence, Table \ref{table: runtime} shows that RSAT is slightly slower than SAM or Riemannian-SAM. In particular, while the RSAT improves the classification accuracies by a notable margin, its wallclock runtime is only slightly slower than SAM by a 1\% gap overall, therefore, it is worth the tradeoff for better performance and robustness.

\begin{table*}[ht]
\centering
\scalebox{1}{
\begin{tblr}{
  colspec = {ccccccc},
  cell{1}{2,6} = {c=2}{c}, 
}
\hline[2pt]
  Method    & \SetCell[c=2]{c} \textbf{CIFAR100} & 2-2
            & \SetCell[c=2]{c} \textbf{CIFAR10} & 2-4
            & \SetCell[c=2]{c} \textbf{AirCraft} & 2-6
            & \SetCell[c=2]{c} \textbf{STL10} & 2-8
   \\
\hline[1pt]
       &    ResNet34 &  ResNet50 &  ResNet34 & ResNet50 &  ResNet34 & ResNet50 & ResNet34 & ResNet50  \\
\cline{2-9}
    SGD    & $18.5_{\pm 1.73}$ & $37.1_{\pm 2.96}$& $18.4_{\pm 1.72}$& $34.8_{\pm 3.05}$& $54.6_{\pm 2.59}$& $111.5_{\pm 4.77}$ & $21.7_{\pm 2.07}$ & $40.1_{\pm 2.06}$& \\

    RSGD    & $27.1_{\pm 2.05}$ & $43.3_{\pm 3.19}$& $25.1_{\pm 1.64}$& $43.6_{\pm 0.99}$& $61.9_{\pm 1.52}$& $127.7_{\pm 2.25}$ & $32.8_{\pm 3.03}$ & $46.1_{\pm 1.99}$&\\
    
    SAM    &   $46.1_{\pm 1.68}$    & $80.9_{\pm 2.79}$  & $45.8_{\pm 1.62}$ & $81.7_{\pm 2.68}$ & $121.3_{\pm 1.3}$ & $240.6_{\pm 4.30}$ & $71.4_{\pm 4.03}$ & $90.2_{\pm 3.22}$   \\
    Riemannian-SAM   & $48.6_{\pm 1.66}$      & $82.7_{\pm 3.14}$ & $46.1_{\pm 1.90}$ & $85.3_{\pm 2.79}$ & $126.9_{\pm 2.1}$ & $251.2_{\pm 3.82}$ & $74.1_{\pm 2.15}$ & $95.9_{\pm 3.34}$\\
    
    RSAT   & $47.1_{\pm 1.03}$      & $84.3_{\pm 1.42}$ & $47.6_{\pm 1.03}$ & $89.4_{\pm 1.63}$ & $129.0_{\pm 1.4}$ & $258.8_{\pm 2.20}$ & $77.0_{\pm 2.01}$ & $97.6_{\pm 4.02}$\\
\hline[2pt]
\end{tblr}
}
\caption{Per-epoch wall-clock runtime in seconds.}
\label{table: runtime}
\vspace*{-\baselineskip}
\end{table*}

\section{Optimization over G := SO(p)}

To solve for the optimal teleportation step $g^*$, we parameterized the orthogonal with the Cayley transform and perform gradient descent. This strategy was widely used by prior works such as \cite{opt} when the optimization problem involves orthogonal constraints. Even though the Cayley transform involves matrix inverse, in practice, this operation is not costly since the size of this matrix is relatively small, which is only $32$ as implemented in our experiment. In this case, we did not observe any significant numerical instabilities due to the relatively small dimensionality. Nevertheless, for larger $p$, one can use Riemannian optimizers,
such as RSGD, to solve for the optimal orthogonal matrix. This is feasible since \( \mathrm{SO}(p) \) is a special case of the Stiefel manifold \( \mathrm{St}(p, p) \). To assess the impact of these different optimization approaches, we compare the Cayley transform and RSGD performance in Table \ref{tab: cayley vs rsgd}. On average, the results indicate \textbf{no} notable differences between the two methods in this setting.

\begin{table}[ht]
\centering

\resizebox{\linewidth}{!}{%
\begin{tabular}{c|ccccc}
\hline
Model    & Method & CIFAR100 & CIFAR10 & STL10 & Aircraft \\ \hline
ResNet34 & Cayley & 80.35    & 96.64   & 89.13 & 87.52    \\
         & RSGD   & 80.22    & 96.67   & 89.31 & 87.64    \\ \hline
ResNet50 & Cayley & 81.52    & 96.88   & 91.72 & 85.52    \\
         & RSGD   & 81.47    & 96.84   & 92.01 & 85.51    \\ \hline
\end{tabular}
}
\caption{Cayley transform vs. Riemannian-SGD to optimize $G^* \in \mathrm{SO}(p)$ on the SupCon loss with $p=32$.}
\label{tab: cayley vs rsgd}
\end{table}

\section{Experimental Details}

\label{appendix: experimental details}
\begin{table}[ht]
\centering
\begin{tabular}{ll}
\toprule
\textbf{Hyperparameters} & \textbf{} \\ 
\midrule
Training epoch                   & 500           \\
Warmup epoch                     & 10               \\
Batch size (CIFAR, STL10, Aircraft)       & 256, 256, 64               \\
Crop size (CIFAR, STL10, Aircraft)       & 32, 64, 224               \\
$\rho$ (SAM, RSAM, and RSAT)         & 0.05             \\
Learning rate                    & 0.5              \\
Learning rate scheduler                    & cosine annealing              \\
Learning rate decay              & 0.1      \\
Weight decay                     & 0.001            \\
Momentum                         & 0.9              \\
Number of Teleport steps         & 5               \\
Teleport period (epochs)         & 20               \\
Size of $\mathbb{U}$      & $d \times 32$ where                \\
\bottomrule
\end{tabular}
\caption{Training hyperparameters of \textsc{RSAT}}
\label{hyper-param}
\end{table}
\paragraph{Experimental Settings}
We conducted our experiments with four datasets for classification tasks, including CIFAR10, CIFAR100, STL10, and Aircraft. The experiment was trained on a signal A100 GPU with 20GB of RAM. More details of the related hyperparameters can be found in Table \ref{hyper-param}.

\paragraph{Teleportation Steps.}
In the teleportation steps, theoretically, we need to find the optimal action $g^*$. However, we do not need to obtain the optimal $g^*$ in practice for an improved performance. Instead, at the teleportation step, we run 5 iterations to solve for $g$. Also, we only teleport after every 20 epochs. These details were also specified in Table \ref{hyper-param}.

\section{Related Works}
\paragraph{Sharpness-Aware Minimization.\quad} Sharpness-aware Minimization (SAM), a technique designed to seek “flat” regions of the loss landscape \citep{foret2021sharpnessaware}, has garnered significant attention for its superior generalization capabilities. SAM has been successfully applied across diverse domains, including multi-task learning, domain generalization, and federated learning, especially in both language and vision models \cite{qu2022generalized, abbas2022sharp, chen2021vision, bahri-etal-2022-sharpness, cha2021swad}. Recent research has focused on enhancing SAM’s performance by investigating its geometric properties in various settings \citep{kwon2021asam, kim22f, zhuang2022surrogate}. Efforts to improve SAM’s efficiency by reducing its computational overhead originating from the two-step gradient calculations have also been pursued \citep{du2022sharpness, liu2022towards}. Comparative studies between SAM and methods like Stochastic Weight Averaging (SWA) \citep{izmailov2018averaging} have provided further insights, with additional connections drawn to Bayesian theory through Bayesian SAM (BSAM) \citep{bsam_iclr23, kaddour2022when}. In Bayesian Neural Networks, sharpness-aware techniques have been extended to posterior distributions \citep{va2023_flat_Bayes}, and the relationship between SAM and distributional robustness has been explored through Optimal Transport methods \citep{va2023_dr_model}.

\paragraph{Optimization on Manifolds.\quad} In machine learning and computer vision, it is often advantageous to impose specific constraints on a model’s parameters, such as symmetric positive definiteness (SPD) \citep{spd}, orthogonality \citep{orthocnn}, or full rank \citep{orthocnn, siamesemanifold, wang2020orthogonal} to improve model robustness. These constraints effectively limit the search space from a general Euclidean space to a more structured manifold, allowing the optimization to leverage intrinsic geometry. Therefore, training can be made more efficient and improve the performance, as the search is confined to a lower-dimensional, relevant subspace \citep{siamesemanifold, book}. This approach has led to the development of manifold-based methods across various applications. For example, in Gaussian mixture models, \cite{spd} proposed a technique that maintains SPD constraints by optimizing on an SPD manifold. To facilitate optimization on such constrained spaces, several Riemannian optimization techniques have been developed, including Riemannian gradient descent (RGD) \citep{rgd}, which enforces manifold constraints but faces computational challenges. To address these, \cite{rsgd} introduced the Riemannian stochastic gradient descent (RSGD) method, which has become widely used on diverse manifolds, including SPD manifolds. More recently, \cite{rsam} proposed Riemannian-SAM, which extends sharpness-aware minimization to the Riemannian setting, improving generalization by refining parameter optimization on the manifold.

\paragraph{Group-Invariant Loss Function.} In deep learning optimization, the loss function is often invariant under the actions of a group $G$. This group invariance allows the model to leverage symmetries which can be particularly beneficial in computer vision tasks. For example, convolutional neural networks exhibit invariance to translations due to their convolutional structure, making them well-suited for image classification \citep{lecun1998gradient, goodfellow2016deep}. Similarly, rotation-invariant models have been designed for tasks where rotational transformations do not affect the target output, such as in medical imaging and satellite image analysis \citep{cohen2016group, esteva2019guide, cheng2017remote}. Various studies have effectively utilized this principle to enhance optimization efficiency and generalization. For instance, \cite{teleportoriginal} introduced the concept of \textit{teleportation}, where parameters are shifted within an orbit to maximize the update step. Furthermore, \cite{siamesemanifold} proposed the qConv layer, which leverages a quotient manifold for improved generalization across different tasks. Theoretical insights from \cite{manifoldbook} also show that Riemannian optimizers, like Riemannian Gradient Descent (RGD), meet the Hessian condition for faster convergence on the quotient space, whereas this condition may fail in the full space. These findings emphasize the theoretical and practical advantages of optimizing adopting the quotient space, especially in vision tasks where symmetries play a critical role.

\section{Riemannian Manifolds}
\label{appendix: riemannian manifolds}
\begin{figure}
\centering
        \includegraphics[width=1\linewidth]{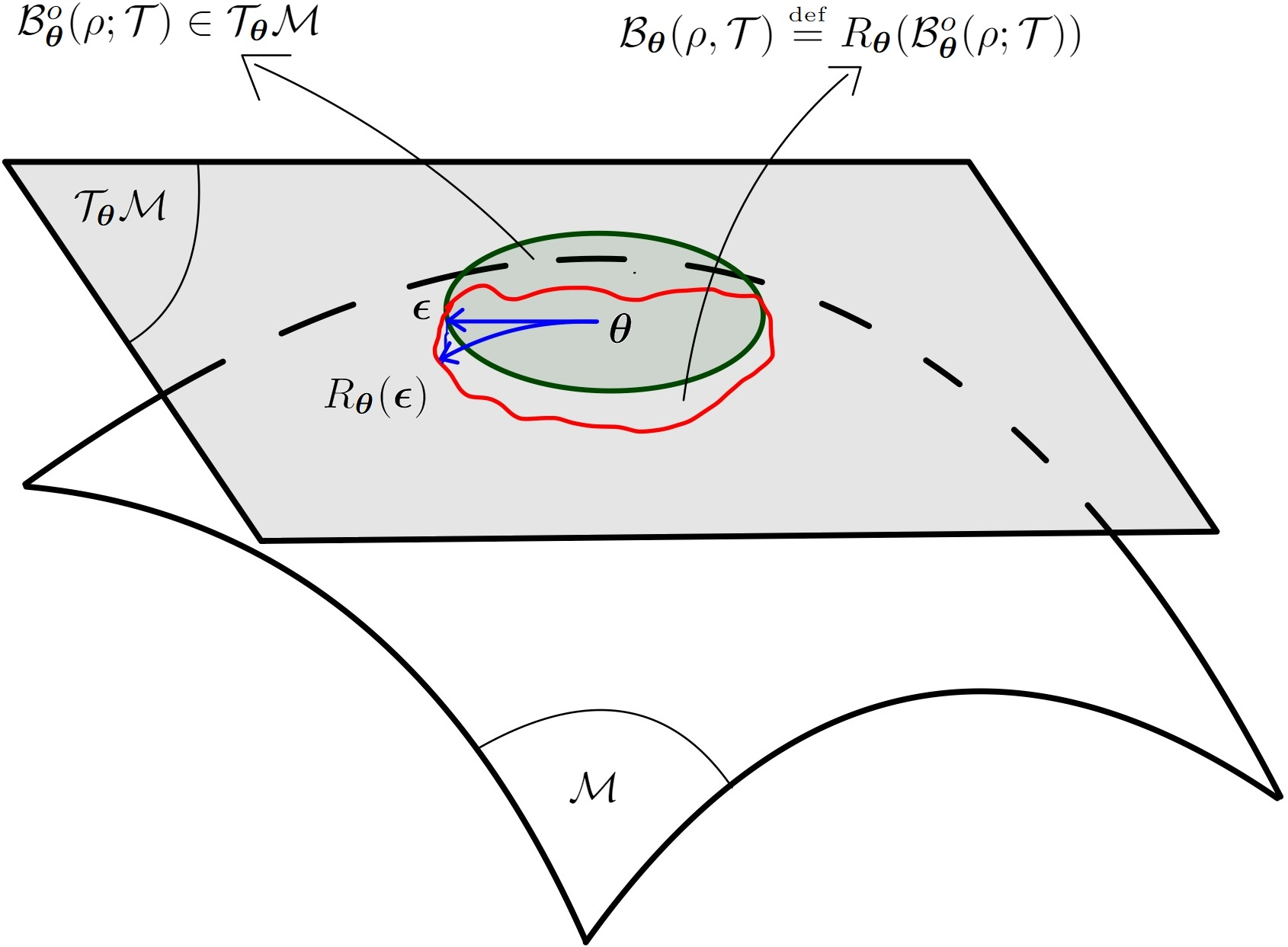}
        \caption{Notations on manifolds. $\mathcal{B}_\theta(\rho; T)$ (green) is an $\rho$-ball on the tangent space $ T_\theta\mathcal{M}$. The neighborhood $\mathcal{B}_\theta(\rho;\mathcal{M})$ (red) of $\theta$ on $\mathcal{M}$ is the retraction image of $\mathcal{B}^o_\theta(\rho; T)$.}
        \label{neighborhood on manifold} 
\end{figure}
This section discusses the key concepts of Riemannian manifolds in detail. Recall that to ensure model robustness, it is typically assumed that some conditions are imposed on the models (e.g., orthogonality, full rank, or SPD constraints), making the model parameters $\mathbf{\theta}$ belong to a \textit{low-dimensional manifold} $\mathcal{M} \subset \mathbb{R}^k$ embedded in the \textit{ambient vector space} $\mathbb{R}^k$, where the dimension $d = \text{dim }\mathcal{M}$ is notably smaller than $k$. Given a $\mathbf{\theta} \in \mathcal{M}$, denote $T_\mathbf{\theta} \mathcal{M}$ as the \textit{tangent space} of $\mathcal{M}$ at $\mathbf{\theta}$ with $\theta$ is the origin of $T_{\mathbf{\theta}}$. Conventionally, $\mathbf{\epsilon} \in T_\mathbf{\theta} \mathcal{M}$ specifies the offset from $\mathbf{\theta}$ on the tangent space, which is the vector $\mathbf{\theta} + \mathbf{\epsilon}$ in the ambient vector space $\mathbb{R}^k$.

The Riemannian manifold is an abstract version of the description above, in which $\mathcal{M}$ is a ''smooth'' manifold. Now we attempt to formally define the concept of \textit{Riemannian manifold}. First consider the union of all tangent space will form a manifold, defined as

\begin{equation*}
    T \mathcal{M} = \{(\mathbf{\theta}, \mathbf{v}): \mathbf{\theta} \in \mathcal{M} \text{ and } \mathbf{v} \in T_\mathbf{\theta} \mathcal{M} \},
\end{equation*}

which is also a manifold. For a point $\theta \in \mathcal{M}$, the tangent space is an Euclidean space, which means we can define the inner product of this space. We denote the inner product on $T_\theta \mathcal{M}$ to be $\langle, \rangle_\theta: T_\theta \mathcal{M} \times T_\theta\mathcal{M}\rightarrow \mathbb{R}$. A metric on $\mathcal{M}$ is a choice of this inner product for each $\theta$. The metric is called Riemannian if for all vector tangent fields 
$V,W:\mathcal{M}\rightarrow T\mathcal{M}$ and $V(\theta),W(\theta)\in T_\theta\mathcal{M}$, the function $\theta\mapsto \langle V(\theta),W(\theta)\rangle_\theta$ is smooth from $\mathcal{M}$ to $\mathbb{R}$. Then, a manifold equipped with a Riemannian metric is called a \textbf{Riemannian manifold}. 

Consider a given pair of $(\theta, v) \in T\mathcal{M}$, there can be many trajectories $c$ on $\mathcal{M}$ starting from $\theta$, following the direction of $v$. This curve can be formulated along a curve $c:[0,1]\rightarrow \mathcal{M}$ such that $c(0) = \theta, c^{\prime}(0) = v$. A\textbf{ retraction operation} on $\mathcal{M}$ is a map $\mathcal{R}:T\mathcal{M}\rightarrow \mathcal{M}$ in which $\mathcal{R}(\theta,tv) = c(t)$. For simplicity, we write $\mathcal{R}_\theta(tv) = c(t)$ for the retraction map starting at $\theta$.

Now we formally define the Riemannian gradient of a smooth map $f: \mathcal{M} \to \mathbb{R}$. We consider a more general case where $f: \mathcal{M} \to \mathcal{M}'$ is a smooth map between two manifolds. For any tangent vector $\mathbf{v} \in T_\mathbf{\theta} \mathcal{M}$, there exists a smooth curve $c$ on $\mathcal{M}$ passing through $\mathbf{\theta}$ with velocity $\mathbf{v}$. Then, the curve $c': t \mapsto f(c(t))$ defines a curve on $\mathcal{M}'$ passing through $f(\mathbf{\theta})$, thus passing through $f(\mathbf{\theta})$ with a certain velocity. This velocity is a tangent vector of $\mathcal{M}'$ at $f(\mathbf{\theta})$ by definition, which we will call the \textit{differential} of $f$ at $\mathbf{\theta}$ along $\mathbf{v}$. Formally, this differential of $f$ at the point $\mathbf{\theta} \in \mathcal{M}$ is defined as the linear map $Df(\mathbf{\theta}): T_\mathbf{\theta} \mathcal{M} \to T_{f(\mathbf{\theta})} \mathcal{M}'$ given by $Df(\mathbf{\theta})[\mathbf{v}] = \frac{d}{dt}f(c(t)) \Big|_{t = 0}$, where $c$ is a smooth curve on $\mathcal{M}$ passing through $\mathbf{\theta}$ at $t = 0$ with velocity $\mathbf{v}$. Now consider the special case where $\mathcal{M}' = \mathbb{R}$, which means $f$ is a smooth, real map, the \textbf{Riemannian gradient} of $f$ is defined as the unique vector field $\text{grad}_\mathbf{\theta} f$ on the tangent space of $\theta$ satisfying: 
\begin{equation*}
    \forall (\mathbf{\theta}, \mathbf{v}) \in T \mathcal{M}; Df(\mathbf{\theta})[\mathbf{v}] = \left<\mathbf{v}, \text{grad}_\mathbf{\theta} f(\mathbf{\theta})\right>_\mathbf{\theta},
\end{equation*}
in a neighbourhood of $\theta$ on $\mathcal{M}$. Finally, the orthogonal projection  $\mathcal{P}_\mathbf{\theta}$  onto $T_\mathbf{\theta} \mathcal{M}$ is defined as: 
\begin{equation*}
    \mathcal{P}_\mathbf{\theta}: \mathbb{R}^k \to T_\mathbf{\theta} \mathcal{M}: \mathbf{w} \mapsto \mathcal{P}_\mathbf{\theta}(\mathbf{w}),
\end{equation*}
with $\left <\mathbf{w} - \mathcal{P}_\mathbf{\theta}(\mathbf{w}), \mathbf{v} \right > = 0$ for all $\mathbf{v} \in T_\mathbf{\theta}\mathcal{M}$. Once an orthogonal basis is chosen for $T_\mathbf{\theta} \mathcal{M}$, $\mathcal{P}_\mathbf{\theta}$ is represented as a (symmetric) matrix. Thus, for readability purposes, we use the notation $\mathcal{P}_\mathbf{\theta} \mathbf{w}$ as a matrix multiplication instead of $\mathcal{P}_\mathbf{\theta}(\mathbf{w})$ as a linear map. These key concepts, including the definition of neighborhoods on manifolds as previously mentioned in the Preliminaries section of the main text, are illustrated in Figure \ref{neighborhood on manifold}.

\section{SPD Manifolds}
\label{appendix: spd manifold}

The summary of the operations on the SPD manifolds is provided in Table \ref{tab: spd operations}.

     \begin{table}[h]
     \centering

\resizebox{\columnwidth}{!}{%
\begin{tabular}{l|ll}
\hline
Manifold & Euclidean $\mathbb{R}^n$ & SPD($n$) with Log-Euclidean metric \\ \hline
Dimension $\mathrm{dim}(\mathcal{M})$        & $n$         & $\frac{n(n+1)}{2}$             \\
Tangent space $\mathcal{T}_x\mathcal{M}$        & $\mathbb{R}^n$         & $\{V|V = V^\top\}$            \\
Projection        & $v$         & $\frac{V+V^\top}{2}$             \\
Exp Map  $\exp_x(v)$      &     $x+v$     &$\exp(\log(X)+V)$\\
 \hline
\end{tabular}
}
\caption{Summary of SPD manifold operations. We provide equivalent operations on Euclidean space
for reference. }
\label{tab: spd operations}
\end{table}

\section{Quotient manifolds and Grassmann manifolds}
\label{appendix: quotient manifolds}

This section briefly discusses the key properties of quotient manifolds and the explicit forms for the Grassmann manifold operations used in our experiments. Only for this section, we use the conventional notations in the literature of differential geometry that $\overline{\mathcal{M}}$ being the total space, while $\mathcal{M} = \overline{\mathcal{M}}/\sim$ denotes the quotient space. Let $\pi: \overline{\mathcal{M}} \to \mathcal{M}$ be the quotient map. When the quotient manifold is Riemannian, we have access to an inner product $\left<\cdot, \cdot\right>_x$ on $T_x\overline{\mathcal{M}}$, which suggests the definition of the horizontal and vertical spaces.

\begin{definition}
    For a quotient manifold $\mathcal{M} = \overline{\mathcal{M}}/\sim$, the \textit{vertical space} at $x \in \overline{\mathcal{M}}$ is the subspace
    \[V_x = \mathsf{Ker}(D\pi(x)) = T_x\mathcal{F},\]
    where $\mathcal{F}$ is the fiber of $x$, $\mathsf{Ker}$ is kernel of linear map. If $\overline{\mathcal{M}}$ is Riemannian, the orthogonal complement of $V_x$ is the horizontal space at $x$:
    \[H_x = (V_x)^\perp = \{u \in T_x \overline{\mathcal{M}}:\langle u,v\rangle_x = 0 \text{ for all }v \in V_x\}.\]
\end{definition}
    Since $D\pi(x) = V_x$, the restricted linear map
    \[D\pi(x)|_{H_x}: H_x \to T_{[x]}\mathcal{M}\]
    is bijective. Via this map, we may define the \textit{horizontal lifts} of a tangent vector of the quotient manifold.
    \begin{definition}
        Consider a point $x \in \mathcal{M}$ and a tangent vector $\xi \in T_{[x]}\mathcal{M}$, the \textit{horizontal lift} of $\xi$ at $x$ is the (unique) horizontal vector $u \in H_x$ such that $D\pi(x)[u] = \xi$. We write
        \begin{align*}u = (D\pi(x)|_{H_x})^{-1}[\xi] = \mathsf{lift}_x(\xi).\end{align*}
    \end{definition}
    Given a retraction $\overline{\mathcal{R}}$ on the total space $\overline{\mathcal{M}}$, we may try to define a retraction $\mathcal{R}$ on the quotient manifold $\mathcal{M}$ as follows: 
    \begin{align*}\mathcal{R}_{[x]}(\xi) = [\overline{\mathcal{R}}(\mathsf{lift}_x(\xi))].\end{align*}
    
    If this is well defined, that is, if the right-hand side does not depend on the choice of lifting point $x \in [x]$, this is indeed a retraction.
    \begin{theorem}
        If the retraction $\overline{\mathcal{R}}$ on the total space $\overline{\mathcal{M}}$ satisfies
        \[x \sim y \implies \overline{\mathcal{R}}_x(\mathsf{lift}_x(\xi)) \sim \overline{\mathcal{R}}(\mathsf{lift}_y(\xi))\]
        for all $x, y \in \overline{\mathcal{M}}$, then $\overline{R}$ defines the retraction on $\mathcal{M}$
    \end{theorem}
    \begin{proof}
        The proof can be found in \textbf{Theorem 9.33} by \cite{manifoldbook}.    
        \end{proof}
Back to the Grassmann manifold, according to \cite{manifoldbook}, the QR retraction mentioned in Section \ref{section: applications} satisfies the property above. Thus, we have the valid retraction on the quotient manifold $\text{St}(n,p)/O(p)$ given by:
\begin{align*}\mathcal{R}^{QR}_{[X]}(\xi) = [\mathsf{qfactor}(X+\mathsf{lift}_X(\xi))],\end{align*}
where $\text{qfactor}$ extracts the Q-factor of a QR decomposition of a matrix in $\mathbb{R}^{n\times p}$. 

We can also define the Riemannian gradient of the quotient manifolds. Indeed, we have the following proposition:
\begin{proposition}
    The Riemannian gradient of $f$ on a Riemannian quotient manifold is related to the Riemannian gradient of the lifted function $\overline{f} = f \circ \pi$ on the total space via
    \begin{align*}\mathsf{lift}_x(\mathrm{grad}f([x])) = \mathrm{grad}\overline{f}(x).\end{align*}
\end{proposition}
\begin{proof}
    The proof is given in Page 235 by \cite{manifoldbook}.
\end{proof}

Thus, the Riemannian gradient of the quotient manifold can be computed from the Riemannian gradient of the ambient space.

\subsection{Advantages of optimizing on quotient manifolds}
\label{appendix: advantages}
This section discusses the advantages of learning on the quotient manifolds. Consider Riemannian Gradient Descent \citep{rgd}, which takes the following update on the total space
\begin{align*}x_{k+1} = \overline{\mathcal{R}}_{x_k}\big(-\alpha_k \text{grad}_{x_k}f(x_k)\big).\end{align*}
 On the other hand, the update on the quotient manifold is given by
\begin{align*}[x_{k+1}] = \mathcal{R}_{[x_k]}\big(-\alpha_k \text{grad}_{[x_k]}f([x_k])\big).\end{align*}

According to \textbf{Section 4.6} by \cite{manifoldbook}, we may expect linear convergence to a local minimizer if the Hessian of the cost function at that point is positive definite. However, the cost function $\overline{f}$ on the total space $\mathcal{M}$ \textbf{cannot} admit such critical points because of its invariance under $\sim$, which is given by the following lemma.
\begin{lemma}
    Let $\overline{\mathcal{M}}$ be a Riemannian manifold and $f: \mathcal{M} \to \mathbb{R}$ be a smooth map on the quotient manifold with the quotient map $\pi$. If $x \in \overline{\mathcal{M}}$ is a critical point for $\overline{f} = f \circ \pi$, then the vertical space $V_x$ is included in $\mathsf{Ker} \mathsf{Hess}\overline{f}(x)$. In particular, if $\text{dim}(\mathcal{M}) < \text{dim}(\overline{\mathcal{M}})$ , then $\mathsf{Hess}\overline{f}(x)$ is not positive definite. 
\end{lemma}
\begin{proof}
    The proof can be found in Page 237 \citep{manifoldbook}.
\end{proof}

According to the lemma, the standard theory does not predict fast local convergence of RGD on $\overline{f}$. Moreover, given a sequence $(x_n)_n$ on $\overline{\mathcal{M}}$, the sequence of equivalence classes $([x_n])_n$ may converge to a limit point in $\mathcal{M}$, while the sequence $(x_n)_n$ \textit{does not} converge in $\mathcal{M}$, hence justify the advantages of quotient out the optimization problem.

\section{Proofs of Theorems and Propositions}

\subsection{Proof: SimCLR loss and SupCon loss are invariant under the actions of SO(p)}
\label{proof: invariant}

This section proves the claim in Section \ref{metric learning} that under the architecture given in Figure \ref{fig: architecture}, the SupCon and SimCLR loss functions are invariant under the right actions of $SO(p)$ on the Grassmannian matrix $U$. Recall that these two loss functions involve the inner product:
\begin{equation*}
    \left<\mathbf{h}_1, \mathbf{h}_2 \right> = \mathbf{h}^{\top}_1\mathbf{M}\mathbf{h}_2 = (\mathbf{h}_1^{\top}\mathbf{U}\mathbf{S} ) \cdot (\mathbf{h}_2^{\top}\mathbf{U}\mathbf{S})^{\top},
\end{equation*}
in which $\textbf{U}$ lies on the Grassmann manifold, and $\textbf{S}$ is a diagonal matrix. Consider $\textbf{U'} = \textbf{UR}$ in which $\textbf{R} \in SO(p)$ is an orthogonal matrix, we have the new matrix $\textbf{M}$ given by:
\begin{align*}
    \mathbf{M'} &= \mathbf{U'S^2U'^\top}\\
    &= \mathbf{URS^2R^\top U^\top} \\
    &= \mathbf{US^2RR^\top U^\top} \\
    &= \mathbf{US^2U^\top = M}. 
\end{align*}
Then, the inner products between the logits remain unchanged, indicating that the loss functions are invariant under the right actions of $SO(p)$ on the matrix $\mathbf{U}$. 

\subsection{Proof of Proposition \ref{proposition: invariant sharpness}}
\label{appendix: proof invariant sharpness}
\begin{proof}

Let $g'$ be any element of $G$. Denote, $\overline{g} = g' \cdot g$. Since $G$ is a group, we also have $\overline{g} \in G$. It follows that:
    \begin{align*}
    \mathfrak{S}(g^{\prime}\cdot\theta; G, M) &= \min_{g \in G} \max_{\theta^{\prime} \in \mathcal{B}_{g\cdot(g^{\prime}\cdot\theta)}(\rho;\mathcal{M})}L_\mathcal{S}(\theta^{\prime})\\
    &=\min_{g\in G}\max_{\theta^{\prime} \in \mathcal{B}_{\overline{g}\cdot\theta}(\rho;\mathcal{M})} L_\mathcal{S}(\theta^{\prime})\\
    &=\min_{\overline{g}\in G}\max_{\theta^{\prime} \in \mathcal{B}_{\overline{g}\cdot\theta}(\rho;\mathcal{M})} L_\mathcal{S}(\theta')\\
    &= \mathfrak{S}(\theta; G, \mathcal{M}).
    \end{align*}
\end{proof}
\subsection{Unpacking the group-invariant sharpness}
\label{appendix: group-invariant sharpness derivation}
We unpack our RSAT loss function into the summation of the original loss function with the group-invariant sharpness term. In particular, we have:
\begin{align*}
    &L^{RSAT}_\mathcal{S}(\theta) = \min_{g \in G}\max_{\theta' \in \mathcal{B}_{g\cdot\theta}(\rho; \mathcal{M})}L_\mathcal{S}(\theta')\\
    &=\min_{g \in G}\max_{\theta' \in \mathcal{R}_{g\cdot\theta}(\mathcal{B}^o_{g\cdot\theta}(\rho; T_{g\cdot\theta}\mathcal{M}))} L_{\mathcal{S}}(\theta') \\
    &=L_{\mathcal{S}}(\theta) + \Big( \underset{:= \mathfrak{S}(\theta; G, \mathcal{M})}{\underbrace{\min_{g \in G}\max_{\|\epsilon\|_2 \leq \rho, \epsilon\in  T_{g\cdot\theta}\mathcal{M}}  L_{\mathcal{S}}(\mathcal{R}_{g\cdot\theta}(\epsilon)) -  L_{\mathcal{S}}(\theta)}}\Big)\\
    &= L_\mathcal{S}(\theta) + \mathfrak{S}(\theta;  G, \mathcal{M}).
\end{align*}

The second term, which is our sharpness term, can be further unpacked as:

\begin{align*}
    &\mathfrak{S}(\theta; G, \mathcal{M})\\
     &=\min_{g \in G}\max_{\epsilon \in \mathcal{B}^o_{g\cdot\theta}(\rho;  T_{g\cdot\theta}\mathcal{M})} \left<\text{grad}_{g\cdot\theta}  L_{\mathcal{S}}(g\cdot\theta), \epsilon \right>_{g\cdot\theta} + \mathcal{O}(\|\epsilon\|^2_\theta)\\
    \label{eq: minimax group-invariant sharpness}
    &\approx \min_{g \in G}\max_{\|\epsilon\| \leq \rho, \epsilon \in T_{g\cdot\theta}\mathcal{M}}\left<\text{grad}_{g\cdot\theta} L_{\mathcal{S}}(g\cdot\theta), \epsilon \right>_{g\cdot\theta},
\end{align*}
where the second inequality was derived from the Taylor expansion on the Riemannian manifolds, which can be found in \cite{intromanifolds}. 

\subsection{Closed form solution to Eq. \ref{eq: minimax group-invariant sharpness}}\label{appendix proof: closed from max}
Firstly, we restate the optimization problem: 

\begin{equation}
  \max_{\|\epsilon\|^2_2\leq \rho^2, \epsilon \in  T_\theta  \mathcal{M} } \big\langle \text{grad}_\theta(L_{\mathcal{S}}(\theta)) , \epsilon\big\rangle_{\theta},
\end{equation}
in which we replaced $g\cdot\theta$ in the original problem with $\theta$ because $G$ acts on $\mathcal{M}$. Let $\text{grad}_\theta L(\theta)^{\top}\mathbf{D}_\theta = \textbf{v}_\theta^{\top}$ and $(\textbf{u}_{\theta,j})$ be an orthonormal basis of  $ T_\theta \mathcal{M}$. Then, it follows that:
\begin{align*}
    \textbf{u}_{\theta,i}^{\top} \textbf{u}_{\theta,j} =  \delta_{i,j}.
\end{align*}
Under the assumption that the $\textbf{u}_{\theta,j}$ form a basis in the tangent space at $\theta$, there exist $\beta_j$ such that
\begin{align*}
\epsilon = \sum_j \beta_j \textbf{v}_{\theta,j}.
\end{align*}
It deduces that
\begin{align*}
\epsilon^{\top}\epsilon = \Big[\sum_j \beta_j \textbf{u}_{\theta,j}\Big]^\top \Big[\sum_j  \beta_j \textbf{u}_{\theta,j}\Big] = \sum_j \beta_j^2.
\end{align*}
We have the Lagrangian objective being:
\begin{align*}
    \textbf{v}_\theta^{\top} \epsilon + \lambda \big[\epsilon^{\top}\epsilon - \rho^2 \big] = \sum_{j}\beta_j \textbf{v}_\theta^{\top}\textbf{u}_{\theta,j} + \lambda \Big[\sum_j \beta_j^2 -\rho^2\Big].
\end{align*}
Taking derivative with respect to $\lambda$ and $\beta_j$, we get the following system of equations:
\begin{align*}
\sum_j \beta_j^2 &= \rho^2\\
\textbf{v}_{\theta}^{\top}\textbf{u}_{\theta,j} + 2\lambda \beta_j &= 0.
\end{align*}
Solving the second equation of the system yields:
\begin{align*}
\label{eq:lambda epsilon}
   \beta_j = -\frac{1}{2\lambda}\textbf{v}_{\theta}^{\top}\textbf{u}_{\theta,j}.
\end{align*}
Substituting into the first equation of the system, we get:
\begin{align*}
&\frac{1}{4\lambda^2}\sum_j\big[\textbf{v}_{\theta}^{\top}\textbf{u}_{\theta,j}\big]^2 = \rho^2\\
\Rightarrow &\frac{1}{2\lambda} = \pm  \rho\Big\{\sum_j\big[\textbf{v}_{\theta}^{\top}\textbf{u}_{\theta,j} \big]^2 \Big\}^{-\frac{1}{2}}.
\end{align*}
Then, the optimal solution to the maximization problem is given by:
\begin{align*}
    \epsilon^* & = \rho\sum_j \frac{\textbf{v}_{\theta}^{\top}\textbf{u}_{\theta,j}}{\sqrt{\sum_j \big[\textbf{u}_{\theta}^{\top}\textbf{u}_{\theta,j} \big]^2}}\textbf{u}_{\theta,j} \\
    &= \rho\sum_j \frac{\text{grad}_\theta L(\theta)^{\top}\mathbf{D}_\theta  \textbf{u}_{\theta,j}}{\sqrt{\sum_j \big[\text{grad}_\theta L(\theta)^{\top}\mathbf{D}_\theta \textbf{u}_{\theta,j} \big]^2}} \textbf{u}_{\theta,j}.
\end{align*}

\subsection{Proof of Proposition \ref{alg: relaxed}}\label{appendix proof: related solution}
Firstly, we restate the optimization problem: 
\begin{equation}
    \label{eq:relaxed max - vepsilon}
   \max_{\|\epsilon\|_2 \leq \rho}\text{grad}_\theta(L_{\mathcal{S}}(\theta))^\top \mathbf{D}_\theta \epsilon 
\end{equation}
     where we replaced $g.\theta$ with $\theta$. We prove that the optimal solution of the problem in Eq. (\ref{eq:relaxed max - vepsilon}) occurs on the boundary. Suppose on the contrary that the optimal solution is $\epsilon^*$ satisfies $(\epsilon^*)^\top  \epsilon^* = (\rho^*)^2 < \rho^2$. Then, $-\text{grad}_\theta L(\theta)^\top \mathbf{D}_\theta \epsilon^* \leq 0$, otherwise we may replace $\epsilon^*$ with $-\epsilon^*$ that still satisfies the constraint and arrive at a strictly smaller objective. However, if we replace $\epsilon^*$ with $\overline{\epsilon} = \epsilon^* \sqrt{\frac{\rho}{\rho^*}}$, it follows: 

    \begin{equation*}
        \overline{\epsilon}^\top \overline{\epsilon}  = \rho^2
    \end{equation*}
and arrive at a smaller objective since $\overline{\epsilon} > \epsilon^*$, which is a contradiction since we are assuming that $\epsilon^*$ is the optimal solution.
    
Therefore, the optimal solution $\epsilon^*$ occurs on the boundary. Thus, the problem is reduced to:  
    \begin{subequations}
    \label{convex prob}
    \begin{align}
        \text{maximize: } \text{grad}_\theta(L(\theta))^\top \mathbf{D}_\theta \epsilon \\%
        \text{subject to: } \epsilon^\top  \epsilon - \rho^2 = 0
    \end{align}
    \end{subequations}
This problem has the Lagrangian: 
\begin{equation*}
    L(\epsilon, \lambda) = -\text{grad}_\theta(L(\theta))^\top \mathbf{D}_\theta \epsilon + \lambda (\epsilon^\top  \epsilon - \rho^2).
\end{equation*}
The stationary point of $L$ satisfies $\frac{\partial L}{\partial \epsilon} = 0$ and $\frac{\partial L}{\partial \lambda} = 0$, which is equivalent to: 

\begin{subequations}
\begin{align*}
    -\text{grad}_\theta(L(\theta))^\top \mathbf{D}_\theta + 2\lambda \epsilon = 0 \\ 
    \epsilon^\top  \epsilon = \rho^2    .
\end{align*}
\end{subequations}
Thus, we have the system of equations: 
\begin{subequations}
\begin{align*}
     \epsilon = \frac{1}{2 \lambda} \text{grad}_\theta(L(\theta))^\top \mathbf{D}_\theta \\ 
    \epsilon^\top \epsilon = \rho^2.
\end{align*}
\end{subequations}
Substituting the first equation into the second one, we get:
\begin{equation*}
    \frac{1}{4\lambda^2}\mathbf{D}_\theta^\top \text{grad}_\theta(L(\theta))  \text{grad}_\theta(L(\theta))^\top \mathbf{D}_\theta = \rho^2,
\end{equation*}
which follows that: 
\begin{align*}
    \frac{1}{2\lambda} & = \frac{\rho}{\Big((\text{grad}_\theta(L(\theta))^\top \mathbf{D}_\theta)^\top  (\text{grad}_\theta(L(\theta))^\top \mathbf{D}_\theta) \Big)^\frac{1}{2}} \\
    &= \frac{\rho}{\left \|\text{grad}_\theta(L(\theta))^\top \mathbf{D}_\theta \right \|_2}.
\end{align*}
Therefore, the maximization problem governs a closed-form solution:

\begin{equation*} \label{eq: ascent vepsilon}
    \overline{\epsilon} = \rho
    \frac{\text{grad}_\theta(L(\theta))^\top \mathbf{D}_\theta}
    {\left \|  \text{grad}_\theta(L(\theta))^\top \mathbf{D}_\theta \right \|_2.
    }
\end{equation*}
Replacing $\theta$ by $g. \theta$, we finish our proof.

\subsection{Proof of theorem \ref{main theorem}}
\label{proof main theorem}
Firstly, we state a lemma whose proof can be found in \cite{manopt_conv_rate}.

\begin{lemma}\label{lemma:Lipschitz_condition}\citep{manopt_conv_rate}
    There exists a constant $C_{ \mathcal{M}} > 0$ such that for any $\theta \in \mathcal{M}$ and $\epsilon \in T_\theta \mathcal{M}$, the following holds:
    \begin{equation*}
        \left\|\mathcal{R}_\theta(\epsilon) - \exp_\theta(\epsilon) \right \|_F \leq C_{ \mathcal{M}}\left \|\epsilon\right\|^2_F.
    \end{equation*}
    \textbf{Remark:} The constant $C_{ \mathcal{M}}$ depends on the manifold structure. Indeed, for retractions on the Stiefel manifold, the constant is independent of $(d, k)$ and can be computed explicitly. Specifically, when using the QR factorization or the polar decomposition as the retraction, we have $C_{ \mathcal{M}} = 1 + \sqrt{2}/2$.
\end{lemma}

Now, we are ready to prove our main theorem. Indeed, we restate the theorem statement:
\setcounter{theorem}{0}
\begin{theorem}
\label{proof: main theorem}
Assuming that the loss function $L$ is $K-$Lipschitz, and the parameter space $\mathcal{M}$ is bounded. Then, for any  small $\rho > 0$ and $\delta \in [0;1]$, with a high probability $1 - \delta$ over training set $\mathcal{S}$ generated
from a distribution $D$, the following holds:
\begin{align*}
    \displaystyle L_D(\theta) & \leq \min_{g\in G}\max_{\theta' \in \mathcal{B}_{g.\theta}(\rho;  \mathcal{M})} L_\mathcal{S}(\theta') \\
    & +  \mathcal{O}\Bigg(C_\mathcal{M}\rho^2 + \sqrt{\frac{d + \log \frac{n}{\delta}}{n-1}}\Bigg).
\end{align*}
\end{theorem}

\begin{proof}
Firstly, we prove that for any $\theta \in \mathcal{M}$, one have: 
\begin{align*}
    &\displaystyle L_D(\theta) \leq \max_{\theta' \in \mathcal{B}_\theta(\rho;  \mathcal{M})} L_\mathcal{S}(\theta') +  \mathcal{O}\Bigg(C_ \mathcal{M}\rho^2 + \sqrt{\frac{d + \log \frac{n}{\delta}}{n-1}}\Bigg).
\end{align*}
Since $\mathcal{M}$ is bounded, it follows that for every $\varepsilon > 0$, there exists a set $\{\theta_i\}_{i = 1}^J$ of predefined points on the manifold $ \mathcal{M}$ that forms an $\varepsilon$-net of $ \mathcal{M}$ with respect to the geodesic distance on $ \mathcal{M}$, which means that for each $\theta \in  \mathcal{M}$, there exists $i$ such that $d_{ \mathcal{M}}(\theta_i,\theta) = d_i < \varepsilon$, in which $d_{ \mathcal{M}}$ is the geodesic distance on manifold $ \mathcal{M}$.

According to the PAC-Bayes generalisation bound of \citep{PACBound}, for any prior distribution $P(\theta)$, with probability at least $1 - \delta$ over the training set $\mathcal{S}$, it holds that

\begin{equation}
    \mathbb{E}_{Q(\theta)}[\ell_\mathcal{D}(\theta)] \leq \mathbb{E}_{Q(\theta)}[\ell_\mathcal{S}(\theta)] + \sqrt{\frac{\text{KL}(Q(\theta)\|P(\theta)+\log \frac{n}{\delta})}{2n-2}}.
    \label{eq: pac bayes 1}
\end{equation}
for any posterior distribution $Q(\theta)$ that may be dependent on the training data $\mathcal{S}$. We define $J$ prior distributions $\{P_j(\theta)\}^J_{j=1}$ as the $J$ Normal laws on the manifold $\mathcal{M}$ centering at $\theta_j$ with the covariance structure $\Sigma = \rho^2 \mathbf{I}$. In particular, the density $p_j(\cdot)$ of $P_j$ is given by:
\begin{equation*}
    p_j(\theta) = k_j\exp(\frac{-1}{2}\|\log_{\theta_j}(\theta)\|^2_{\Sigma^{-1}}),
\end{equation*}

where $k_j^{-1} = \mathcal{O}(\rho^d)+ Vol(\mathcal{M})$We choose $Q(\theta)$ as the distribution on $\mathcal{M}$ with the density:
\begin{equation*}
    q(\theta) = k_0 \exp(\frac{-1}{2}\|\log_{\theta_0}(\theta)\|^2_{\Sigma^{-1}}),
\end{equation*}
which is a Normal law centered at $\theta_0$. One can choose $\theta_0$ arbitrarily from $\mathcal{M}$, and it can be dependent on $\mathcal{S}$. To minimize the bound in Eq. \ref{eq: pac bayes 1}, we aim to choose the prior that is closest to $Q(\theta)$ in terms of the KL divergence. Applying the PAC-Bayes bound in Eq. \ref{eq: pac bayes 1} for each $j$ makes the following hold for each $j$ with probability $1 - \delta_j$ over the choice of the training set $\mathcal{S}$:

\begin{equation}
    \mathbb{E}_{Q(\theta)}[\ell_\mathcal{D}(\theta)] \leq \mathbb{E}_{Q(\theta)}[\ell_\mathcal{S}(\theta)] + \sqrt{\frac{\text{KL}(Q(\theta)\|P_j(\theta)+\log \frac{n}{\delta_j})}{2n-2}}.
    \label{eq: pac bayes 2}
\end{equation}

By having the intersection of the training sets for which Eq. \ref{eq: pac bayes 2}, it implies that Eq. \ref{eq: pac bayes 2} holds for all $P_j$ over the intersection. By the union bound theorem, the probability over the choice of the intersection is at least $1 - \sum^J_{j=1} \delta_j$. By letting $\delta_j = \frac{\delta}{J}$, we derive that:

\begin{align}
    \mathbb{E}_{Q(\theta)}[\ell_\mathcal{D}(\theta)] &\leq \mathbb{E}_{Q(\theta)}[\ell_\mathcal{S}(\theta)]\\
    &+ \sqrt{\frac{\text{KL}(Q(\theta)\|p(\theta)+\log \frac{n}{\delta} + \log J)}{2n-2}}.
\end{align}

Now we choose the prior $P_j(\theta)$ that is as close to the posterior $Q(\theta)$ as possible (in KL divergence). Consider the KL divergence terms: 

\begin{align*}
    &KL(Q\|P_j)=\mathbb{E}_{P_j}[\log P_j - \log Q]\\
    &= \frac{1}{2}\mathbb{E}_{P_j}[d\mathcal{O\log(\rho)}-\frac{1}{\rho}(\|\log_{\theta_j}(\theta)\|^2_2+\|\log_{\theta_0}(\theta)\|^2_2)].
\end{align*}

Consider the second term, we write $\log_{\theta_j}(\theta) = \epsilon_j$. Then, the condition $\
theta \sim P_j$ is equivalent to $\epsilon_j \sim \mathcal{N}(0, \rho^2 \mathbf{I})$, which implies that
\begin{equation*}
    \mathbb{E}_{P_j}[\|\log_{\theta_j}(\theta) \|^2_2] = \mathbb{E}_{\epsilon \sim \mathcal{N}(0, \rho^2\mathbf{I})}[\|\epsilon\|^2] = d\rho.
\end{equation*}



On the other hand, we have $\|\log_{\theta_0}(\theta)\|^2 \leq d_\mathcal{M}(\theta_0, \theta)$, then we have 

\begin{equation*}
    \mathbb{E}_{P_j}[\|\log_{\theta_0}(\theta)\|^2_2] \leq d_\mathcal{M}(\theta_j, \theta_0)^2.
\end{equation*}

Combining these results together, and consider $j^*$ such that $d_\mathcal{M}(\theta_0, \theta_{j^*}) \leq \varepsilon$, we have

\begin{equation}
    KL(Q\|P_j) \leq \frac{1}{2}(\frac{\varepsilon^2}{\rho^2} + d \mathcal{O}(1)).
\end{equation}

Therefore, we have: 
\begin{align*}
    &\mathbb{E}_{\theta\sim\mathcal{N}_{\mathcal{M}}(\theta_0, \rho^2I)}[L_\mathcal{D}(\theta)]\leq \mathbb{E}_{\theta\sim\mathcal{N}_{\mathcal{M}}(\theta_0, \rho^2I)}[L_\mathcal{S}(\theta)] \\
    &+\sqrt{\frac{\frac{\varepsilon^2}{2\rho^2}+\rho \mathcal{O}(1) + \log\frac{n}{\delta}+\log J}{n-2}},
\end{align*}

where $\mathcal{N}_\mathcal{M}$ denotes the Normal law on $\mathcal{M}$. We can rephrase $\mathbb{E}_{\theta\sim\mathcal{N}_{\mathcal{M}}(\theta_0, \rho^2I)}[f(\theta)]$ as $\mathbb{E}_{\epsilon\sim \mathcal{N}(0, \rho^2 G(\theta)^{-1})}f(\exp_{\theta_0}(\epsilon))$, where $G(\theta)$ is the local chart at $\theta$, and replacing $\theta_0$ with $\theta$, it follows that:

\begin{align*}
    &\mathbb{E}_{\epsilon\sim\mathcal{N}(\theta, \rho^2G(\theta)^{-1})}[L_\mathcal{D}(\exp_\theta(\epsilon))]\\
    &\leq \mathbb{E}_{\epsilon\sim\mathcal{N}(\theta, \rho^2G(\theta)^{-1})}[L_\mathcal{S}(\exp_\theta(\epsilon))] \\
    &+\sqrt{\frac{\frac{\varepsilon^2}{2\rho^2}+\rho \mathcal{O}(1) + \log\frac{n}{\delta}+\log J}{n-2}},
\end{align*}

Under the assumption that perturbation does not improve the population loss, we have
\begin{equation*}
    L_\mathcal{D}[\theta] \leq \mathbb{E}_{\epsilon\sim\mathcal{N}(\theta, \rho^2G(\theta)^{-1})}[L_\mathcal{D}(\exp_\theta(\epsilon))].
\end{equation*}

The next step is to bound the expectation in RHS by the worst-case loss. Similar to \cite{sam}, we make use the following result:

\begin{equation*}
    z \sim \mathcal{N}(0, \rho^2 \mathbf{I}) \implies \|z\|^2 \leq k\rho^2(1+\sqrt{\frac{\log n}{k}})^2
\end{equation*}

with probability at least $1 - \frac{1}{\sqrt{n}}$. By letting $u = G(\theta)^{\frac{1}{2}}\epsilon$, we have $u \sim \mathcal{N}(0, \rho^2 \mathbf{I})$. Applying the inequality above, we have with probability at least $1 - \frac{1}{\sqrt{n}}$ that:

\begin{equation*}
\|u\|^2 = \epsilon^\top G(\theta) \epsilon \leq d \rho^2(1+\sqrt{(\log n)/d})^2    
\end{equation*}
Let $\gamma = \rho(\sqrt{d}+\sqrt{\log n})$. Then, we bound the term of interest by partitioning the $\epsilon$ space into those with $\epsilon G(\theta) \epsilon \geq \gamma^2$, and those that are $\epsilon G(\theta) \epsilon \leq \gamma^2$. Implying that 

\begin{align*}
    &\mathbb{E}_{\epsilon\sim\mathcal{N}(\theta, \rho^2G(\theta)^{-1})}[L_\mathcal{S}(\exp_\theta(\epsilon))]\\
    &\leq \max_{\epsilon^\top G(\theta) \epsilon \leq \gamma^2}L_\mathcal{S}(\exp_\theta(\epsilon))+\frac{\ell_{\max}}{\sqrt{n}}\\
    &\leq \max_{\| \epsilon\|_{T_\theta \mathcal{M}}\leq \gamma^2}L_\mathcal{S}(\exp_\theta(\epsilon))+\frac{\ell_{\max}}{\sqrt{n}}.
\end{align*}

Since the loss is $K$-Lipschitz, according to the Lemma \ref{lemma:Lipschitz_condition}, we have 

\begin{align*}
   \big| L_\mathcal{S}(\mathcal{R}_\theta(\epsilon) - L_{D}(\exp_\theta(\epsilon))\big|\leq KC_\mathcal{M}\rho^2.
\end{align*}

Combining these results together, we have: 
\begin{align*}
    &L_\mathcal{D}(\theta) \leq \max_{\theta' \in \mathcal{B}_{\theta}(\rho;\mathcal{M})}L_\mathcal{S}(\theta')+KC_\mathcal{M}\rho^2 + \frac{\ell_{\max}}{\sqrt{n}}\\
    &+\sqrt{\frac{\frac{\varepsilon^2(\sqrt{d}+\sqrt{\log n})^2}{2\gamma^2}+\log \frac{n}{\delta}+\log J + d\mathcal{O}(1)}{2n-2}}\\
    &= \max_{\theta' \in \mathcal{B}_{\theta}(\rho;\mathcal{M})}L_\mathcal{S}(\theta') + \mathcal{O}\Bigg(C_\mathcal{M}\rho^2 + \sqrt{\frac{d+\log\frac{n}{\delta}}{n-1}}\Bigg).
\end{align*}

Recall the condition that $L$ is invariant under the group action. Applying the previous inequality for $g.\theta$, we have that for all $g\in G$:
\begin{align*}
    L_\mathcal{D}(\theta) & = L_\mathcal{D}(g.\theta) \\
    & \leq \max_{\theta^{\prime}\in \mathcal{B}_{g.\theta}(\rho, \mathcal{M})} L_\mathcal{S}(\theta^{\prime}) + K C_{ \mathcal{M}}(\varepsilon + \rho)^2 + K\varepsilon \\ & + \sqrt{\frac{\mathcal{O}(d + \ln(n/\delta))}{n-1}}.
\end{align*}
This is true for all $\varepsilon > 0$. Then, we conclude that:
\begin{align*}
    L_\mathcal{D}(\theta) & = L_\mathcal{D}(g.\theta) \\
    & \leq \max_{\theta^{\prime}\in \mathcal{B}_{g.\theta}(\rho, \mathcal{M})} L_\mathcal{S}(\theta^{\prime}) + K C_{ \mathcal{M}}\rho^2\\ & + \sqrt{\frac{\mathcal{O}(d + \ln(n/\delta))}{n-1}}
\end{align*}
which finishes our proof.
\end{proof}

\end{document}